\documentclass[review]{elsarticle}

\usepackage{lineno}
\modulolinenumbers[5]

\journal{Journal of \LaTeX\ Templates}

\usepackage{amsmath,amssymb,amsfonts}
\usepackage[ruled]{algorithm2e}
\usepackage{graphicx}
\usepackage{textcomp}
\usepackage[flushleft]{threeparttable}
\usepackage[utf8]{inputenc}
\usepackage{booktabs}
\usepackage{float}
\usepackage{csquotes}
\usepackage{times}
\usepackage{url}
\usepackage{array,color}
\usepackage{multirow}
\usepackage{soul}
\usepackage[table]{xcolor}
\usepackage{lastpage,fancyhdr,graphicx}
\usepackage{epstopdf}
\usepackage{rotating}
\usepackage{bm}
\usepackage[colorinlistoftodos]{todonotes}
\usepackage{epstopdf}
\usepackage{color,soul}
\usepackage{caption}
\usepackage[caption=false]{subfig}

\usepackage[utf8]{inputenc}

\SetCommentSty{mycommfont}
\usepackage{xr}
\usepackage[colorlinks]{hyperref}
\begin{document}
\begin{frontmatter}

\title{Towards Interpretable ANNs: An Exact Transformation to Multi-Class Multivariate Decision Trees}

\author[mymainaddress]{Duy T. Nguyen\corref{mycorrespondingauthor}}
\cortext[mycorrespondingauthor]{Corresponding author}
\ead{tung.nguyen@student.adfa.edu.au}

\author[mymainaddress]{Kathryn E. Kasmarik}

\author[mymainaddress]{Hussein A. Abbass}

\address[mymainaddress]{Trusted Autonomy Laboratory, School of Engineering and Information Technology, University of New South Wales - Canberra, Canberra 2600, Australia}

\begin{abstract}\label{abstract}

On the one hand, artificial neural networks (ANNs) are commonly labelled as black-boxes, lacking interpretability; an issue that hinders human understanding of ANNs\textquoteright \ behaviors. A need exists to generate a meaningful sequential logic of the ANN for interpreting a production process of a specific output. On the other hand, decision trees exhibit better interpretability and expressive power due to their representation language and the existence of efficient algorithms to transform the trees into rules. However, growing a decision tree based on the available data could produce larger than necessary trees or trees that do not generalise well. In this paper, we introduce two novel multivariate decision tree (MDT) algorithms for rule extraction from ANNs: an Exact-Convertible Decision Tree (EC-DT) and an Extended C-Net algorithm. They both transform a neural network with Rectified Linear Unit activation functions into a representative tree, which can further be used to extract multivariate rules for reasoning. While the EC-DT translates an ANN in a layer-wise manner to represent exactly the decision boundaries implicitly learned by the hidden layers of the network, the Extended C-Net combines the decompositional approach from EC-DT with a C5 tree learning algorithm to form decision rules. The results suggest that while EC-DT is superior in preserving the structure and the fidelity of ANN, Extended C-Net generates the most compact and highly effective trees from ANN. Both proposed MDT algorithms generate rules including combinations of multiple attributes for precise interpretations for decision-making.

\end{abstract}


\end{frontmatter}


\section{Introduction}\label{section-1}

Symbolic AI techniques adopt interpretable and explainable representation languages with sufficient expressive power for a human to understand the system\textquoteright s behaviours~\cite{shortliffe1984model,johnson1994agents,swartout1991explanations,van2004explainable}. Nevertheless, symbolic systems that rely on deductive logic lack the ability to adapt to changes in the environment and context, require excessive knowledge of the problem domain in advance, and often end up with a hard-to-maintain knowledge base. 

Inductive learning affords a machine with an ability to form and update its own internal representation using experiences. Data labelled with ground truth, environmental (including user) feedback, or self-guidance using pre-designed similarity metrics guide the three basic forms of learning: supervised, reinforcement and unsupervised learning, respectively. 

Early machine learning and statistical inferencing algorithms required significant feature engineering efforts by the human designer. This effort substantially decreased with the arrival of deep learning algorithms~\cite{wen2016discriminative,wu2016enhanced,salakhutdinov2013learning}. 
With the possibility of implementing an artificial neural network (ANN) on a hardware, and the ability to leverage the vectorisation available in graphics processing units (GPUs) in software, ANNs offer significant advantages when it comes to speed of learning and ability to handle big data. Recent advances in black-box models such as deep ANNs have shown significant opportunities to learn and even surpass human-level performance in some tasks~\cite{mnih2015human,churchland2016computational,mnih2016asynchronous}. 
However, one of the major disadvantages normally associated with ANNs is their lack of interpretability; that is, they are labelled as black-boxes. The highly distributed nature of their implicit capture of knowledge and their ability to approximate highly non-linear functions using many local units and different layers of transformations, complicate the interpretability of ANNs. In the absence of appropriate representations to interpret what an ANN learns, social integration, human acceptance, verification against requirements or previous knowledge bases, and performance assurance are some of the main challenges to the use of ANNs in practical applications, especially in safety-critical environments. Addressing the drawback of these opaque systems to enable seamless and safe interactions between humans and machines can be attained by developing new techniques for explainable models without sacrificing the performance of the AI models.

Interpretability is desirable throughout the training and decision-making processes of a model so that users can have a better understanding of that model, which is one of the key factors to establish human trust in artificial intelligence. Although poorly defined in literature, the term `interpretation' can be simply described as a process involving the transformation of a type of information into another, which provides a better understanding for a specific group of users. Different studies refer to two categories of the subject including algorithmic interpretability and model interpretability~\cite{hepworth2020human}. The former is related to the methods investigating the mechanism by which an algorithm works, and assist in debugging the system by revealing the knowledge of the algorithm\textquoteright s structure and its system of parameters. The latter one concerns only the model mapping the inputs to outputs. In this paper, the `interpretability' term refers to the first category. For a model that has not learnt the task well-enough, interpretability could shed light on which part of the model is lagging behind. For a model that generalises well, interpretability sheds light on the rationale behind the model\textquoteright s behaviour. Among the techniques to interpret the hidden layers of the ANNs, decision trees~\cite{tsujino1995implementation,craven1996comprehensible,Schmitz1999ANN-DT,hayashi2010understanding,chakraborty2019rule} have been used to learn the relationship between the input and output of a learned neural network. This provides a means to extract and represent the implicit knowledge in the network in an interpretable form. 

In this paper, we propose two multivariate decision tree algorithms called the Exact-Convertible Decision Tree (EC-DT) and the Extended C-Net algorithm. Extended C-Net learns the relationship between the last hidden layer and the outputs, then infers the input-output relationships through recursive back-projections guided by the ANN\textquoteright s weights. EC-DT constructs the rule sets layer-wise based on the activation of the hidden nodes to find a precise representation of ANNs in a tree form. To analyze the ability of each method in interpreting the neural network, we use three indicators: fidelity of the rules as a measure of how well the rules matches the network they were extracted from, the number of rules and premises/constraints in each rule as a measure of the size of the interpretation model, and the level of transparency of the model\textquoteright s interpretation. 

The contributions of the paper are three-fold. Firstly, the paper introduces novel multivariate tree algorithms as decompositional rule extraction techniques for inspecting  ANNs with a focus on continuous input and Rectified Linear Unit (ReLU) activation function. Secondly, a deeper investigation is conducted on the relationship between rule set compactness and complexity, and the complexity of the data space to address the question of which situations a simple black-box (pedagogical) rule extraction technique are more favored, taking into account the trade-off required between the fidelity and transparency. Thirdly, the paper presents some methods to transform the decision rules into a more interpretable type of representation to bridge the gap between a mathematical interpretation and a  common-sense one.

The organization of this paper is as follows. Section~\ref{section-2} reviews the literature including previous decision tree approaches and multivariate decision trees (MDTs) used to extract rules from ANNs. Section~\ref{section-3} introduces the EC-DT algorithm, which can convert an ANN to representative rules while preserving 100$\%$ of the ANN\textquoteright s performance.  Section~\ref{section-4} describes the Extended C-Net algorithm to learn the decision rules from the ANN models using back-projection techniques.  Experimental setup and the set of metrics we use to assess the performance and the interpretability of our approach are then presented in Section~\ref{section-5}. We then discuss our results and corresponding analysis in Section~\ref{section-6} and conclude the paper in Section~\ref{section-7}.

\section{Background}\label{section-2}

In this section, we present a short review of the recent literature on ANN interpretation methods, followed by the requirements and various types of rule extraction techniques for ANNs. We then present related work on several multi-variate decision tree methods to extract interpretable representations from ANNs.

\subsection{Interpreting Black-box Models}\label{section-2.1}

ANNs have demonstrated a significant social impact due to their universal function approximation properties, robustness, very large scale implementation characteristic, generalization abilities, and success in many applications. However, due to their black-box nature, they are not as widely acceptable by humans when compared to classic rule-based systems that rely on symbolic representations~\cite{saad2007neural,alexander1999template}. Opaque models like ANNs need to explain their decision-making processes, to be transparent without sacrificing their predictive power. 

Currently, the explainable artificial intelligence (XAI) domain calls for solutions to overcome the opaqueness of ANNs to improve their reliability and trustworthiness when they are used in decision support engines and expert systems. Approaches in XAI might alleviate problems of knowledge extraction in these black-box systems, provide the systems with explanation and reasoning abilities, facilitate the verification and validation of the model, and inspect and diagnose the sources of erroneous interferences~\cite{andrews1995survey,taha1999symbolic}. These abilities could enhance the utility, transparency, and explainability of ANN in safety critical applications~\cite{gunning2017explainable,adadi2018peeking,samek2019explainable}. Techniques in the XAI domain revolve around two main approaches: \textit{transparent machine learning models} and \textit{post-hoc explanation}~\cite{markus2020role}. While the former produces intrinsically interpretable models that are ready to implement the decision-making processes in a form that can be reasoned with by the users, post-hoc methods attempt to explain the original decision-making models whose criteria of simulatability, decomposability and algorithmic transparency are not met, especially when the models are not simple enough or are used to represent complex non-linearity~\cite{arrieta2020explainable}.

Beside the transparent machine learning approaches, recent literature for interpreting ANNs focuses on three approaches: \textit{model-specific}, \textit{model-agnostic} and \textit{example-based explanation} methods~\cite{daudt2021research}. Model-specific methods are often designed for specific types of neural network models. Deep visualization methods, such as DevNet, Grad-CAM, and Grad-CAM++, produce saliency maps from the convolutional layers of the Convolutional ANNs (CNNs) whose activated areas serve as the regions with highest correlation to the models\textquoteright \ decisions or the objects that need to be identified in the input images~\cite{gan2015devnet,selvaraju2017grad,8354201}. The advantage of these methods is to provide an understanding of how different convolutional layers\textquoteright \  features are formed  in  response  to  the  input  images. Despite the fact that this technique might be beneficial for domain experts, visualization of a saliency map alone does not provide a sound representation in a form that could be understood by a wide variety of users. These methods can be further extended to include an explanation structure, such as a Recurrent Neural Network (RNN), for both visual and language description. Hendrick et al.~\cite{hendricks2016generating} advance a visual explanation model from captioning models to take into account both image-relevant and class-relevant properties. A fine-grained CNN is used to identify components in given images and link to appropriate linguistic terms, while multiple Long-Short Term Memory (LSTM) layers generate a sequence of words that fuses all details into an explanation. The explanations provided by this type of models are much more intuitive due to the existence of both visual marking and explanatory language. Other model-specific algorithms are interpretable CNN-based methods~\cite{zhang2018interpretable}, knowledge graph-based methods~\cite{wang2019explainable,ma2019jointly} and feature importance based on activation propagation~\cite{shrikumar2017learning}. These techniques are currently most suited for specific applications and specific models. 

Model-agnostic methods can be designed to interpret other black-box models, normally in a post-hoc fashion. Due to the fact that this type of models does not bind or play as an intrinsic part of the prediction models, it is more flexible in design concept and does not affect the performance of decision-making processes. Model-agnostic can be divided into two main categories called feature importance, and knowledge extraction~\cite{adadi2018peeking,markus2020role,islam2021explainable,confalonieri2021historical}. Feature importance methods compute and rank the contributions of the input attributes to the outputs of the models. Some examples include Individual Conditional Expectation (ICE) and Shapely values, which have been applied to explain the decision-making processes in classification and control problems~\cite{goldstein2015peeking,casalicchio2018visualizing,fidel2020explainability,nascita2021xai,zhang2020explainable}.

In this paper, we focus on the model-agnostic methods for knowledge extraction approaches, which extract the knowledge governing the mapping between the model\textquoteright s inputs and outputs without modifying the original operation of the networks~\cite{li2020survey}. They generate surrogate models that can be translated easily into a human-understandable language. Extracting knowledge from ANNs can be implemented at local or global scale and has been practiced in many applications such as classifying products in industries~\cite{amin2013novel} as well as business analysis~\cite{hayashi2010understanding}.

Local explanation takes advantages of non-complex interpretable models such as decision trees to estimate the relationships between the inputs and outputs of  opaque models such as ANNs, through a learning process with a subset of training data representing specific sub-area of the problem space~\cite{guidotti2018local}. Local interpretable model-agnostic explanations (LIME) algorithm~\cite{ribeiro2016should} generates local interpretations by mapping a specific input instance to its corresponding output. The expanding of such mapping to a local area is performed by creating a cluster of perturbation process which randomly samples some data close to the original one. Although LIME can create local approximation of the input-output relationships from a black-box model, the simplicity of the local algorithm may produce low-fidelity interpretations when applied to highly non-linear sub-regions~\cite{guidotti2018local}. This challenge is alleviated in a later study by an algorithm, called Anchor, which adapts a high-fidelity set of rules that generalizes the local approximations~\cite{ribeiro2018anchors}. Multiple generated rules in Anchor can overlay the local sub-regions determined by the original LIME process to a high precision given a pre-defined confidence level. Recently, there are several other methods trying to reduce the issue of instability of the original LIME algorithm such as MeLIME~\cite{botari2020melime}, OptiLIME~\cite{visani2020optilime}, and s-LIME~\cite{zhou2021slime}, or modify it for different applications~\cite{haunschmid2020audiolime,rabold2020enriching}. LIME, s-LIME, and Anchor are later used as baseline methods for comparison with our proposed algorithms.

Global knowledge extraction consists of approaches that learn a model that can interpret the entire decision-making processes of a black-box model. Popular methods in this category include decision trees or decision rules. We will expand on some global rule extraction algorithms below.

\subsection{Global Rule Extraction from Artificial ANNs}\label{section-2-criteria}

The usefulness of a rule extraction algorithm depends on multiple criteria. Taha et al.~\cite{taha1999symbolic} listed different aspects that need to be considered when designing a rule extraction system from ANNs including: 
\begin{itemize}
    \item \textbf{Level of detail}: The presentation of information in the explanation based on hypotheses of the system.
    \item \textbf{Comprehensiveness:} The fidelity of rules to represent the knowledge within the black-box system.
    \item \textbf{Comprehensibility:} The property of rule set to identify knowledge of a model's processes.
    \item \textbf{Transparency:} The ability of the rules to be easily inspected in order to inform explanations or decisions.
    \item \textbf{Generalization:} The performance of the extracted rule set on new, unprecedentedly observed data.
    \item \textbf{Mobility:} The ability to apply the rule extraction algorithm to different network architectures.
    \item \textbf{Adaptability:} The ability to modify the set of extracted rules when the networks are updated after a further training session.
    \item \textbf{Theory refinement:} The ability to overcome restrictions due to missing data or inaccurate domain knowledge.
    \item \textbf{Robustness:} The insensitivity of the extracted rules to noise in the training data.
    \item \textbf{Computational complexity:} Demand of computational resources for extracting the set of rules and the execution of inference relative to the size and number of attributes in the dataset.
    \item \textbf{Scalability}: The ability of the rule extraction algorithm to scale relative to the change in problem complexity and network structure.
\end{itemize}

The list of criteria that was introduced by Taha et al.~\cite{taha1999symbolic} is a general list of criteria for rule extraction algorithms regardless of the purpose of the model. The purpose of the neural network model is to learn the task. As such, criteria such as generalization and robustness are important. For rule extraction algorithms to generate alternative models to ANNs, the use of generalization and robustness are thus important to evaluate them. In the scope of this paper, rule models for interpreting trained ANNs are investigated. Therefore, criteria such as fidelity is important to ensure that the resultant interpretation is a faithful (correct) representation of the original model. In this paper, we investigate the criteria of \emph{comprehensiveness}, \emph{comprehensibility}, and \emph{transparency} which affect the \textbf{performance}, \textbf{compactness} and \textbf{interpretability} of different rule extraction techniques, respectively. The \emph{comprehensiveness} of the rule extraction methods can be represented by the fidelity metrics. The higher the fidelity, the higher the faithfulness of the rule sets in covering the same regions covered by the original ANN model. The \emph{comprehensibility} has a strong relationship with fidelity and the number of extracted rules and premises in each rule expression. The more rules and premises in the rule set, the closer the interpretations to the actual decision-making processes of the ANNs. These can adversely affect the compactness and interpretability of the model as the dimensions will be higher so that it is more difficult for users to understand the interpretation. If one uses a more compressed form of rule representations, the transparency of the interpretation can be improved as the complexity is less. However, the enhancement of the transparency does not necessarily improve the interpretability and usefulness of the rule set as it might have a lower fidelity. Due to the inter-dependency of different criteria, we select  \emph{fidelity}, \emph{number of rules and premises/constraints in each rule}, and \emph{transparency} as the criteria to assess the quality of interpretations. We also look at the computational complexity of different rule extraction algorithms.

\begin{figure}[!htb]
    \centering
    \includegraphics[width=0.6\linewidth]{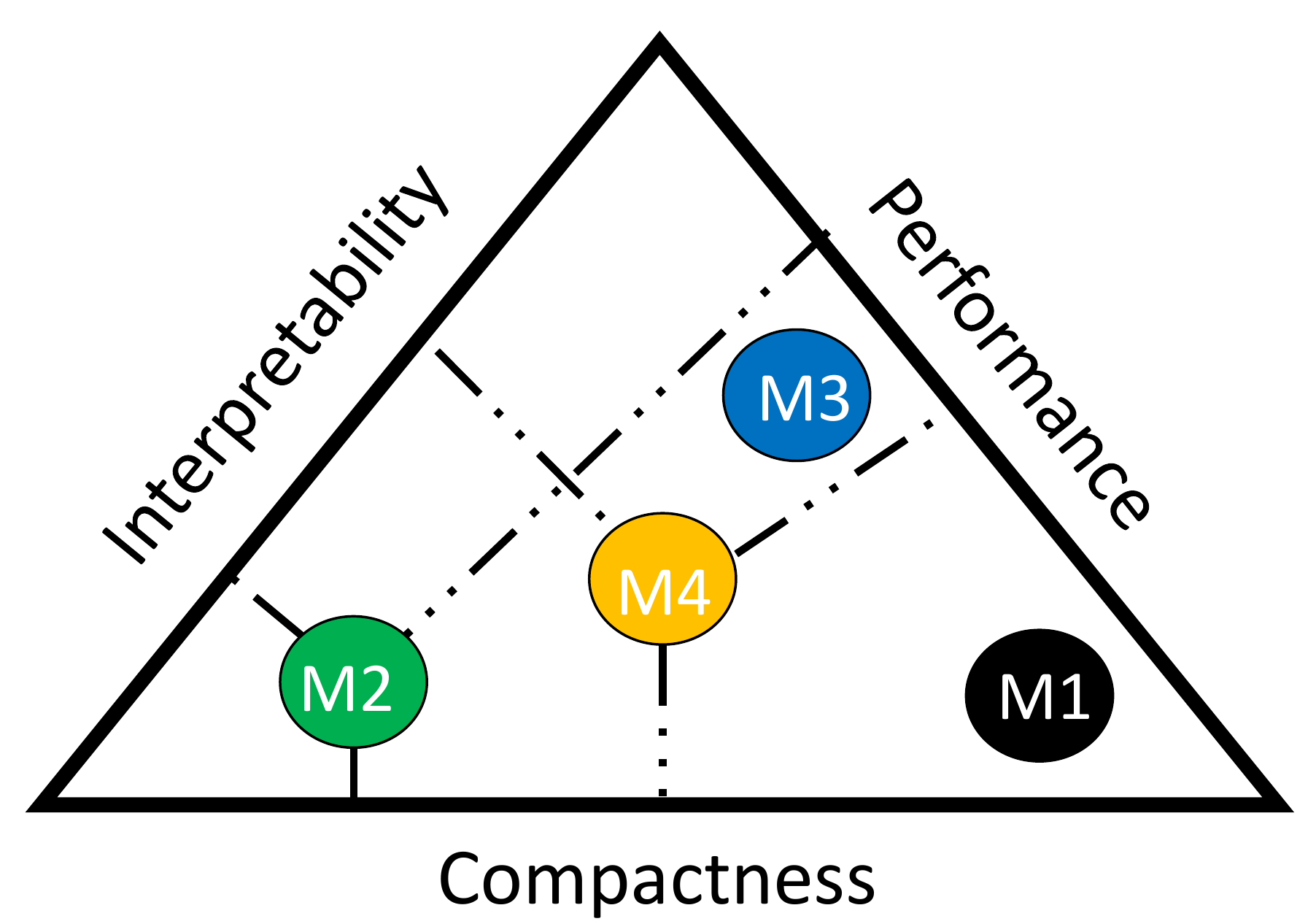}
    \caption{Illustration of interdependent characteristics of a model. Different models are represented by different circles. The closer the distance between a model an an edge of the triangular, the higher the value of the corresponding characteristic and vice versa.}
    \label{fig:three-dimensions-of-models}
\end{figure}

Figure~\ref{fig:three-dimensions-of-models} describes an inter-dependency between three properties of a model including \textit{performance}, \textit{compactness}, and \textit{interpretability}. Different models possess different profiles of these three dimensions. For example, M1 can be a representative of a neural network, whose performance on a task can be relatively high with a very compact representation while its opaqueness causes a very low interpretability. M4 is a high-fidelity rule extraction method that enhances interpretability with easy-to-read rule representations, which may cause a decline in the model\textquoteright s compactness and performance. The trade-off between those properties should consider the primary objectives of analysis. For example, in a human-autonomy interaction situation, where the user needs to understand the rationale behind an action of the autonomous system, it is more important to prioritize interpretability over compactness. On the other hand, in situations like search-and-rescue missions where speed is more important, compactness and performance of the model could be more prominent than interpretability.

Various categories of rule extraction algorithms are reported in the literature. One of the earliest classification frameworks was based on different features that rule extraction methods exhibit including: (1) the expressive power of the extracted rules, (2) transparency of the rule extraction method, (3) usage of specialized training scheme, (4) quality of the extracted rules, and (5) computational complexity~\cite{andrews1995survey}. However, this categorization system seems complex due to the overlap between their elements, such as the strong dependency between the expressive power and quality of extracted rules. Another taxonomy, called \emph{Input-Network-Training-Output-Extraction-Knowledge}, divided clearly the techniques based on modules of the classification frameworks \cite{gupta1999generalized}. Following this type of taxonomy, it is simpler for system designers to select or design the rule extraction algorithms with clear requirements for each component. 

In this paper, we will follow Hruschka et al.~\cite{hruschka2006extracting}\textquoteright s classification due to its popularity and wide acceptance. They divide rule extraction techniques into three categories: \textbf{pedagogical (black-box rule extraction)}, \textbf{decompositional (link rule extraction)}, and \textbf{eclectic (hybrid)} techniques. 

\emph{Pedagogical} approaches are data-driven and only find the direct mapping between the inputs and outputs of the ANN using some machine learning techniques. This set of methods does not reach inside the black-box to find the real links within the ANNs. Quinlan\textquoteright s C4.5~\cite{quinlan1987generating} is one of the most popular algorithms for building a tree representation that utilizes a discrimination process over different data attributes to maximize the information gain ratio. The C4.5 decision tree is commonly used for extracting rules from ANNs. Decision Detection by Rule Extraction (DEDEC) is another rule extraction technique that ranks the input attributes of the input data relative to the outputs of the ANN~\cite{tickle1996dedec}. These rankings are then used to cluster the input space and produce a set of binary rules describing the relationships between the data attributes in each cluster and the ANN\textquoteright s outputs. Other notable examples of black-box rule extraction approaches are BRAINNE~\cite{sestito1992automated}, Rule-extraction-as-learning~\cite{craven1994using}, TREPAN~\cite{craven1996extracting} and BIO-RE~\cite{taha1999symbolic}. The strengths of these approaches lie in their abilities to offer a fast, simple rule set with high transparency and scalability. However, the extracted rules might not be comprehensive and or able to generalize to the test data in various domains.

\emph{Decompositional} rule extraction techniques consider the links between layers of an ANN (the weights and activation at each hidden and output nodes) to compose the rules. These approaches generally describe more accurately the input-output relationship than \emph{pedagogical} approaches. Most techniques in this class consist of two main stages including searching for the weighted sum of the input links for the activation of each hidden node and then producing a rule with inputs as premises. Typical examples of this class of techniques are SUBSET~\cite{towell1993extracting}, KT~\cite{fu1994rule}, NeuroRule~\cite{setiono1996symbolic}, NeuroLinear~\cite{setiono1997neurolinear}, rule extraction by successive regularization~\cite{ishikawa2000rule}, and Greedy Rule Generation~\cite{odajima2008greedy}. Towell et al.~\cite{towell1993extracting} developed a well-known rule-extraction algorithm called MofN that can address the limits of SUBSET in terms of binary inputs, scalability, and repetition of extracted rules. The algorithm achieved a higher fidelity compared to some other black-box and link rule extraction methods. RuleNet~\cite{mcmillan1991connectionist} and RULEX~\cite{andrews2002rule} are decompositional techniques that were specialized for ANNs with localized hidden units. While RULEX extracts rules from a Constrained Error Back-Propagation (CEBP) network whose hidden nodes are localized in a bounded area of the training samples, RuleNet extracts binary rules from a mixture of experts trained on a localized ANN. The extracted rules using these approaches are much more comprehensive, but complicated if the number of attributes or input nodes of the ANN is large. The decompositional approaches face some challenges in transparency, computational complexity, and scalability when applied to large networks.

The analysis in the rest of the paper looks at the inter-dependency between the performance, compactness and interpretability, as discussed above. We compare these metrics of four rule extraction methods. C5.0 decision trees (Pedagogical) and the state-of-the-art techniques called LIME, Stabilized-LIME (s-LIME), and Anchor are selected as baseline methods. Our proposed EC-DT exactly converts a neural network into a decision tree (Decompositional) and the Extended C-Net algorithm is an \emph{eclectic/hybrid} method combining the strengths of both methods above. Classic methods such as the C5.0 decision tree algorithm generates simple set of univariate rules with high interpretability. However, theoretically, this algorithm requires an infinite number of rules to represent oblique decision hyperplanes. As a result, such algorithm might have a low fidelity when capturing the mapping of a neural network. Our assumption is that our proposed models can take advantage of multivariate representations so that they gain a better level of rule compactness and higher performance in complex problems. The EC-DT can preserve the performance better while Extended C-Net has a more balanced profile on all dimensions.

\subsection{Multi-variate Decision Trees}\label{section-2.2}

Decision Trees are interpretable representations able to approximate the underlying functions that ANNs represent. Decision Trees~\cite{boz2002extracting,johansson2009evolving} can approximate the input-output relationship of a neural network and can get converted to a set of decision tables and rules. 

Univariate decision trees are limited to produce a set of rules each operating on a single variable at a time. They rely on axis-parallel decision hyperplanes to approximate the decision boundary, which results in a huge model when the decision boundary is oblique and/or highly nonlinear~\cite{brodley1995multivariate}. These drawbacks limit their capability to produce succinct explanations.

A multivariate decision tree can generate a rule expression in terms of a combination of multiple variables as inputs.
OC1~\cite{murthy1994system} is one of the earliest multivariate decision tree. OC1 searches for optimal set of weights on all data attributes and attempts different weighted sums of input attributes to find the best local decision boundaries. Sok et al.~\cite{sok2016multivariate} modify a univariate alternating decision tree algorithm into a multivariate one by proposing three approaches to weight multiple attributes and replace the base of univariate conditions by combinations of multivariate ones. The resultant trees demonstrate significantly better accuracy while maintaining acceptable fidelity in comparison with their univariate counterparts and ensembles of univariate decision trees. A PCA-partitioned multivariate decision tree algorithm is proposed to solve the multi-label classification problems in large scale datasets~\cite{wang2018efficient}. The multivariate trees, constructed by using the maximum eigenvalue of PCA to choose the weights of each variable in combination, produce a high accuracy.

In the next sections, we describe our proposed multivariate decision tree built with the decompositional rule extraction and hybrid techniques to solve binary and multi-class classification problems.

\section{Transforming an ANN to a Multivariate Decision Tree Using EC-DT Algorithm}\label{section-3}

In this section, we propose a multivariate tree algorithm that transforms an ANN into an equivalent decision tree. This \emph{decompositional extraction} algorithm is designed to be complete and sound; that is, it preserves 100\% performance of ANNs (high fidelity) while providing an interpretable representation.

\begin{figure}[!ht]
\centering
\includegraphics[width=1.0\linewidth]{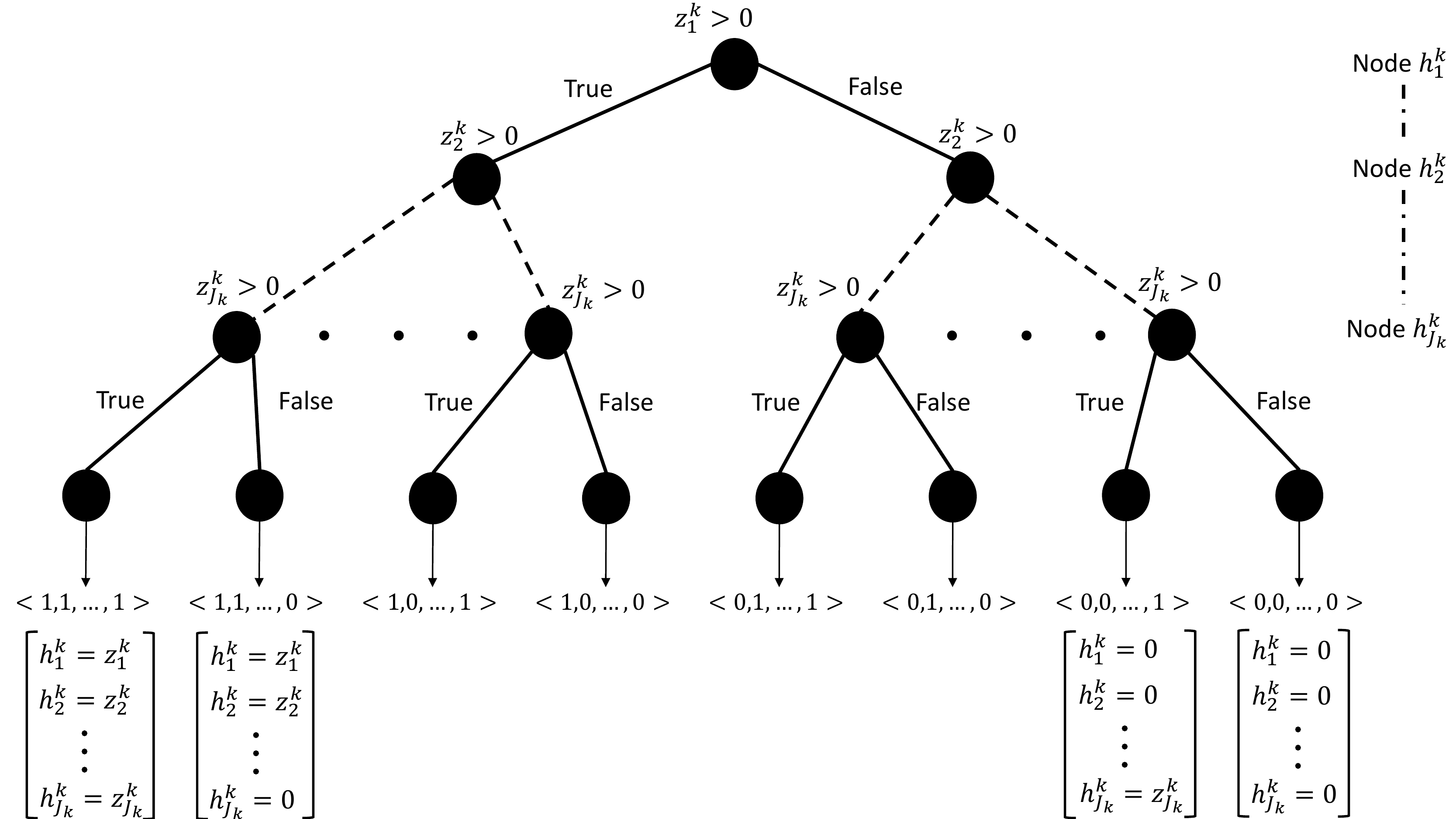}
\caption{Binary EC-DT representing the output of a hidden layer with ReLU activation function.}
\label{fig:extactTree}
\end{figure}

In our study, the activation functions of hidden layers are Rectified Linear Unit (ReLU). The ReLU function has the form:
\begin{equation}
\sigma(f) = \begin{cases} 0, & \mbox{for } f\leq 0 \\ f, & \mbox{for } f > 0 \end{cases}
\label{eq:relu}
\end{equation}

For ReLU, a node is considered activated (i.e. the rule fired) if its output is greater than 0. We might construct a decision tree to represent the constraint by which the activation of each node in the hidden layers is decided. Denote $z_{j_k}$ as the value of hidden node ${j_k}$ in hidden layer $k$ before applying the activation function, that is
\begin{equation}
z^{k}_{j_k} = \sum_{j_{k-1}=1}^{J_{k-1}} w_{j_{k-1} j_k} H^{k-1}_{j_{k-1}} + \beta^k_{j_k}
\end{equation}

For nodes in the first hidden layer, we have 
\begin{equation}
z^{1}_{j_1} = \sum_{i=1}^{I} w_{i j_1} x_i + \beta^1_{j_1}
\end{equation}

The size of the tree grows exponentially as the network size grows; a drawback we will fix below.

Due to the nature of ReLU activation as described in Equation~\ref{eq:relu}, the output of a node after applying this activation is either 0 or the same value to the input to that node, which is a weighted sum of the outputs from previous nodes connected to it. Thus, each hidden layer of the neural network can be transformed into a binary decision tree with each tree stage deciding the activation of the corresponding node in the hidden layer based on the constraint of whether or not the value before the activation function is greater than 0.

\begin{algorithm}[!htbp]
\footnotesize
\SetKwInOut{Input}{Input}
\SetKwInOut{Output}{Output}
\Input{Number of nodes in hidden layers $\{J_1,J_2,...,J_K\}$}
\Output{Decision tree's set of nodes $\mathcal{T}$}
\SetKwInOut{Output}{Initialize}
\Output{$\mathcal{T}\gets\varnothing$, current set of parent nodes $\mathcal{P}\gets\varnothing$, current set of child nodes $\mathcal{C}\gets\varnothing$, node tuple $q^{ID_\tau}(id,hidden\_layer,hidden\_node\_id,$\\$parent\_id,branch,leaf)$} 

Set tree node id $ID_\tau=1$ \\
Set hidden layer $k=1$. \\
Set hidden node id $ID_k=1$. \\
Create root node $q^{r}(ID_\tau,k,ID_k,None,None,leaf=False)$.\\
Add $q^{r}$ to $\mathcal{T}$, add $q^{r}.id$ to $\mathcal{P}$.\\
$ID_\tau\gets2$; $ID_k\gets2$.\\

\While{$k \leq K$}
{
    \While{$ID_k \leq J_k$}
    {
        Set $\mathcal{E}$ the number of elements in $\mathcal{P}$.\\
        \For{$i = 1$ \KwTo $\mathcal{E}$}
        {
            \uIf{$k=K$ and $ID_k=J_K$} 
            {  \tcp{leaves nodes} 
               Create node \tcp*{true branch} $q^{ID_\tau}_1(ID_\tau,k,ID_k,ID_{p_i \in \mathcal{P}},1,True)$ \\
               Create node \tcp*{false branch} $q^{ID_\tau}_0( ID_\tau,k,ID_k,ID_{p_i \in \mathcal{P}},0,True)$ \\
               Find set of branches $S^{ID_\tau}_1$ and $S^{ID_\tau}_0$ leading to $q^{ID_\tau}_1$ and $q^{ID_\tau}_0$ by tracing back to root.\\
               $q^{ID_\tau}_1.value \gets S^{ID_\tau}_1$; $q^{ID_\tau}_0.value \gets S^{ID_\tau}_0$ \\
               $ID_\tau \gets ID_\tau +1$
            }
            \Else
            {
                Create node \tcp*{true branch} $q^{ID_\tau}_1(ID_\tau,k,ID_k,ID_{p_i \in \mathcal{P}},1,False)$ \\
                Create node \tcp*{false branch} $q^{ID_\tau}_0( ID_\tau,k,ID_k,ID_{p_i \in \mathcal{P}},0,False)$ \\
                $ID_\tau \gets ID_\tau +1$
            }
            Add nodes to $\mathcal{T}$.\\
            Add nodes IDs to $\mathcal{C}$.\\
        }
        $\mathcal{P} \gets \mathcal{C}$; $\mathcal{C} \gets \varnothing$. \\
        $ID_k \gets ID_k +1$.
    }
    $k \gets k+1$.
}
\textbf{return} $\mathcal{T}$.
\caption{EC-DT Tree Generation Algorithm}\label{Alg:Build_ECDT}
\end{algorithm}

Given a hidden layer of $J_K$ nodes with ReLU activation, the direct translation of this layer to a binary tree can be illustrated in Figure~\ref{fig:extactTree}. The tree can be considered a multi-target tree with the output being a binary vector of the active nodes (value of 1) and disabled node (value of 0). In the case that a node is activated (satisfying constraint for TRUE branch), the real output of the node after activation is the same as its value prior activation: $h^k_{J_k} = z^k_{J_k}$. That is, we can alternatively replace this activation array by a vector of regression values for each node based on weights.

Algorithm~\ref{Alg:Build_ECDT} illustrates the steps to build an EC-DT from a Neural Network with ReLU activation function. Each leaf node in the produced tree has a list of true or false branches that starts from the root node all the way to the leaf. As a result, we can produce a list of ANN layers\textquoteright \ activations $S=\{S^1,S^2,...S^K\}$, where $S^k$ is an array representing the activations of hidden nodes in layer $k$ of the ANN. The set of rules can be extracted from EC-DT by converting every leaf node in the tree into constraints given the list of hidden nodes\textquoteright activations, the weights matrices, and the biases matrices from ANN (see Algorithm~\ref{Alg:RuleExtraction_ECDT}). The values of weights and biases of disabled nodes do not contribute to the constraints of the next neural layer. Equivalently, we set the weights of connections from the disabled nodes of the current layer to zero and then compute the updated weights and biases matrices representing the linear combination between the input variables and the outputs of the subsequent neural layer. A rule is invalid when it contains two or more conditions that generate a contradiction. The set of rules extracted with this algorithm originally contains rules which are invalid. A rule can be represented by a set of constraints/inequalities. If the constraints in a rule conflict with one another, the rule is invalid and gets eliminated to reduce the size of the set of rules.

\begin{figure}[!ht]
    \centering
    \includegraphics[width=0.8\textwidth]{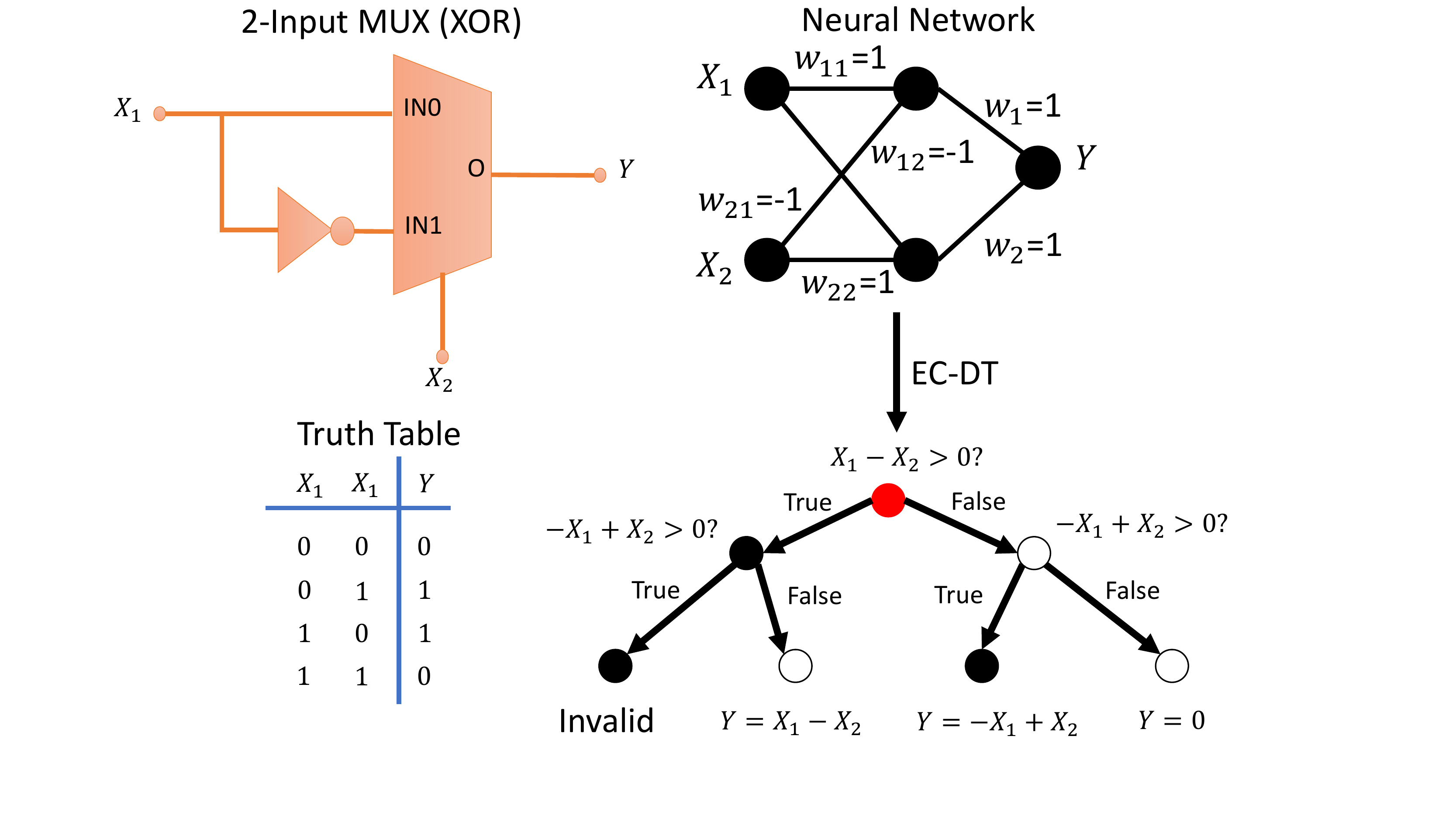}
    \caption{A XOR gate example with binary inputs that can be represented by a neural network which is converted into an EC-DT.}
    \label{fig:MUX-example}
\end{figure}

We present a multiplexer problem with binary input as an example to illustrate how the proposed EC-DT algorithm converts an ANN into a decision tree. Figure~\ref{fig:MUX-example} introduces a gate that takes two binary inputs $X_1$ and $X_2$ and produces a function $Y=XOR(X_1,X2)$. Assume that we have a neural network with weights as shown in Figure~\ref{fig:MUX-example} and zero biases to represent this function. The EC-DT algorithm converts the neural network into a binary tree. Each layer in the tree includes nodes which represent the constraint for testing whether a corresponding unit in the ANN is activated or not. The paths from root to leafs represent combinations of activation of hidden units. This results in four cases (with one invalid case) of output value $Y$. The constraints in each layer and the consequences at the leaves form the rule set of the ANN.

\begin{algorithm}[!htbp]
\small
\SetKwInOut{Input}{Input}
\SetKwInOut{Output}{Output}
\Input{A leaf node with its list of branches corresponding to activations of ANN\textquoteright hidden layers $S=\{S^1,S^2,...S^K\}$, number of nodes in hidden layers $\{J_1,J_2,...,J_K\}$, a set of weights matrices of ANN $\mathcal{W}=\{\mathcal{W}^{I1},\mathcal{W}^{12},...,\mathcal{W}^{(K-1)K},\mathcal{W}^{KY}\}$, and a set of biases matrices of ANN $\mathcal{B}=\{\mathcal{B}^{1},\mathcal{B}^{2},...,\mathcal{B}^{K},\mathcal{B}^{Y}\}$}
\Output{A rule/set of constraints and consequences $\mathcal{R}$}
Set $\mathcal{R} \gets \varnothing$. \\
\For{$k=1$ \KwTo $K$}
{
    \For{$s=1$ \KwTo $J_k$}
    {
        \uIf{$S^k(s)=0$}
        {  
           Convert matrix form $X\mathcal{W}^{Ik}_{*,s}+\mathcal{B}^{Ik}(s)>0$ into linear inequation form. \\  
           Elements in $s^{th}$ row of $\mathcal{W}^{k(k+1)}$  are set to $0$.
        }
        \Else
        {
           Convert matrix form $X\mathcal{W}^{Ik}_{*,s}+\mathcal{B}^{Ik}(s)\leq 0$ into linear inequation form. 
        }
        Add inequation to $\mathcal{R}$ as constraint.
    }
    \uIf{$k=K$}
    {
        Compute $\mathcal{W}^{IY} = \mathcal{W}^{IK}\mathcal{W}^{KY}$. \\
        Compute $\mathcal{B}^{IY} =  \mathcal{B}^{IK}\mathcal{W}^{KY}+\mathcal{B}^{Y}$.    
    }
    \Else
    {
        Compute $\mathcal{W}^{I(k+1)} = \mathcal{W}^{Ik}\mathcal{W}^{(k(k+1)}$. \\
        Compute $\mathcal{B}^{I(k+1)} =  \mathcal{B}^{Ik}\mathcal{W}^{k(k+1)}+\mathcal{B}^{k+1}$.
    }
}
Convert matrix form $\mathcal{Y}=X\mathcal{W}^{IY}+\mathcal{B}^{IY}$ to equations. \\
Add equations to $\mathcal{R}$ as consequences.

\textbf{return} $\mathcal{R}$.
\caption{EC-DT Rule Extraction Algorithm (for a leaf)}\label{Alg:RuleExtraction_ECDT}
\end{algorithm}

\section{Extended C-Net: An approach to learn multivariate decision trees from a neural network}\label{section-4}

Decision trees have been used with ANNs to build an interpretable model that explains the decision-making process of neural-networks. The C-Net algorithm, proposed by Abbass et  al~\cite{abbass2001c}, is one of those early algorithms which uses a univariate decision tree (UDT) to generate a multivariate decision tree (MDT) from ANNs. In this paper, we propose a modification of the algorithm into an extended version of C-Net to extract rules from ANNs. We describe the original algorithm and the modification below, where the Extended C-Net Algorithm is adopted to the specific use of the ReLU activation functions in the hidden layers.

After the ANN is trained, new data set are introduced and the outputs of the last hidden layer are computed. We may split the new data set into training and testing sets for the purpose of training the decision tree. Therefore, from a set of training and test data, denoted as $\langle\mathbf{X}_{training},\mathbf{O}_{training}\rangle$ and $\langle\mathbf{X}_{testing},\mathbf{O}_{testing}\rangle$ respectively, we can compute the mapping between the last hidden output layer and the output, denoted as $\langle\mathbf{H}^K_{training},\mathbf{O}_{training}\rangle$ and $\langle\mathbf{H}^K_{testing},\mathbf{O}_{testing}\rangle$.

We use the data representing the relationship between the last hidden layer and the output of the network to train a C5 decision tree whose algorithm adapts an entropic information gain ratio for branch-splitting criterion and has been demonstrated to be more accurate, and less memory intensive~\cite{quinlan2004data}.  

\subsection{Back Projection of Extended C-Net}\label{section-4.1}

\begin{figure}[!ht]
    \centering
    \includegraphics[width=0.5\textwidth]{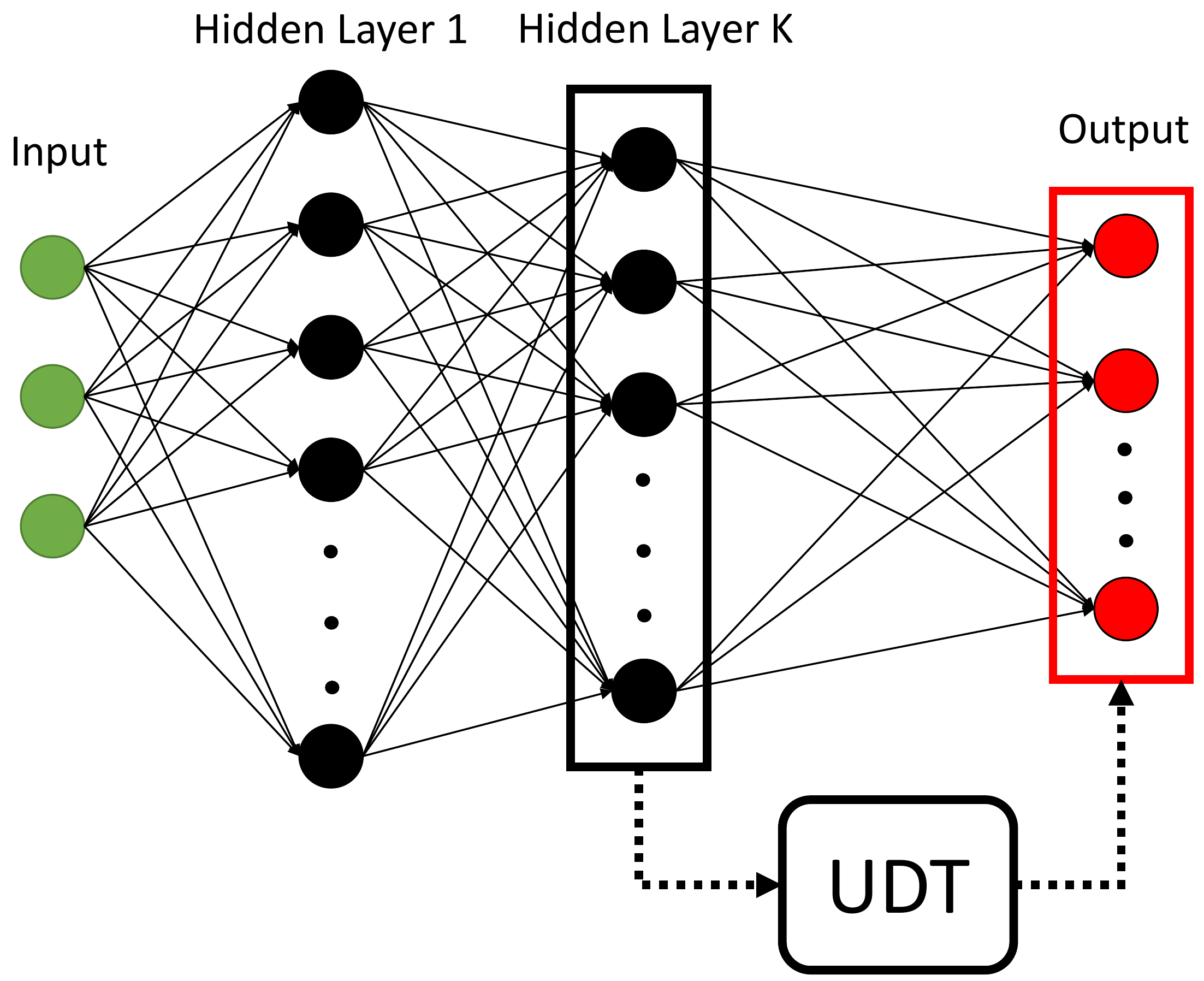}
    \caption{The Extended C-Net framework where C5 algorithm extracts UDT rules between the last hidden layer and the output.}
    \label{fig:Extended C-Net}
\end{figure}

After the ANN is trained, we use the output of the last hidden layer $\mathbf{H}^K(\mathbf{X^t})$ and the class prediction $\mathbf{O}^t$ to be the input and output of the UDT layers, respectively. Figure~\ref{fig:Extended C-Net} illustrates the Extended C-Net architecture.

Commonly, a DT can be represented by a set of polyhedrons expressed in the form of linear constraints. Every constraint as learned by the DT has the form of $H_{j_{K}}(\mathbf{X^t})$ $op$ $C_{j_{K}}$, in which $op$ represents the binary operators $\{\leq,<,=,>,\geq\}$, and $C_{j_{K}}$ is the numeric threshold of such a constraint on input $H_{j_{K}}(\mathbf{X^t})$. It is needed to back-project from the output of the neural network to the input of the neural network to obtain a multivariate form of the expression. Extended C-Net adopts the inverse of the used activation function, that is
\begin{equation}
(\sum_{j_{K-1} = 1}^{J_{K-1}} w_{j_{K-1} j_{K}}H^{K-1}_{j_{K-1}}(\mathbf{X^t})+\beta^K_{j_K}) \;\;\; op. \;\;\; \sigma_K^{-1}(C_{j_{K}})
\end{equation}
In our ANN, the activation functions of hidden layers are ReLU. We chose ReLU due to their linear simplicity and popularity within ANN. Hence the inverse of this function is
\begin{equation}
\sigma^{-1}(f) = \begin{cases} f, & \mbox{for } f>0 \\ \eta, & \mbox{for } f \leq 0 \end{cases}
\end{equation}
where $\eta$ is an arbitrary negative number. ReLU is partially invertible; hence, it is a challenge to find the mathematical expression relating the output back to the input. Some of the nodes might not be activated resulting in a value of zero, which then might not appear in the mathematical combination expression for nodes of the next level. The expression is dynamic at the output depending on the value of inputs.

The input-output mapping is represented by the following expression in case of two hidden layers activated by ReLU, in which the outputs of the second hidden layer are fed as inputs to the UDT:
\begin{equation}
\sigma(\sum_{j_1=1}^{J_1} w_{j_1 j_2}\sigma(\sum_{i = 1}^I w_{ij_1}x_i + \beta^1_{j_1}) + \beta^2_{j_2}) \;\;\;\; op. \;\;\;\; C_{j_2}
\end{equation}
A node is considered active if its output is greater than 0. We could rewrite the non-linear constraint into one linear constraint on the activation function level, and a second linear constraint on an index of the former. We note that a constraint on the index is mathematically too complex, but we will compensate that by using this constraint as a filter. 
\begin{equation}
\sum_{j_1}\sum_{i = 1}^I w_{j_1 j_2}w_{ij_1}x_i + w_{j_1 j_2}\beta^1_{j_1} + \beta^2_{j_2} \;\;\;\; op. \;\;\;\; C_{j_2} 
\end{equation}
rewritten as 
\begin{equation}
\sum_{j_1}\sum_{i = 1}^I w_{j_1 j_2}w_{ij_1}x_i  \;\;\;\; op. \;\;\;\; C_{j_2} - (w_{j_1 j_2}\beta^1_{j_1}) + \beta^2_{j_2}
\end{equation}
with additional constraints:

$\forall j_1$ satisfying  $\sum_{i=1}^I w_{ij_1}x_i + \beta^1_{j_1} > 0$ \\

The multivariate decision tree Extended C-Net is a rewrite of the ANN into interpretable logic using nodes/conditions and branches/information-flows. Each rule induced by each leaf of the resultant Extended C-Net tree can be traversed back to weights of the network to deduce the input-target relationships as represented by the network and its weights. The Extended C-Net algorithm takes advantage of the EC-DT tree generation algorithm, but with leaves produced earlier at layer $K-1$ and target values of those leaves are linear combinations of input $X$ as hidden nodes' outputs at layer $K$. The Extended C-Net algorithm can be described as follows:
\begin{itemize}
    \item \textbf{Step 1:} Feed inputs to ANN and compute the values at final hidden layer's nodes $H^K_{j_K}$.
    \item \textbf{Step 2:} Use $H^K_{j_K}$ values and labels of instances to train a UDT (C5 decision trees) and extract rules from UDT tree.
    \item \textbf{Step 3:} Build a decision tree using algorithm~\ref{Alg:Build_C-Net} (Supplementary Document, Section~\ref{section-S1}) until hidden layer $K-1$.
    \item \textbf{Step 4:} Use Algorithm~\ref{Alg:RuleExtraction_CNet} to extract final set of rules for every leaf.
\end{itemize}

Due to the use of both a data-driven method to build UDT tree and a decompositional method to build the decision tree that represent the constraints between inputs and the last hidden layers, the Extended C-Net algorithm belongs to the \emph{eclectic/hybrid} approaches. An example of how the Extended C-Net algorithm works can be found in Section~\ref{section-S1} in the Supplementary document.

\begin{algorithm}[!htbp]
\small
\SetKwInOut{Input}{Input}
\SetKwInOut{Output}{Output}
\Input{A leaf node with its list of branches corresponding to activations of ANN\textquoteright hidden layers $S=\{S^1,S^2,...S^{(K-1)}\}$, number of nodes in hidden layers $\{J_1,J_2,...,J_{(K-1)}\}$, a set of weights matrices of ANN $\mathcal{W}=\{\mathcal{W}^{I1},\mathcal{W}^{12},...,\mathcal{W}^{(K-1)K}\}$, a set of biases matrices of ANN $\mathcal{B}=\{\mathcal{B}^{1},\mathcal{B}^{2},...,\mathcal{B}^{K}\}$, and a set of constraints from UDT's leaf in form: $\mathbf{H^K_{j_K} \; op \; C_{j_K}} \;\; \forall j_K \in \Upsilon; \;\; \Upsilon \subseteq \{1,2,...,J_K\}$}
\Output{A rule/set of constraints and consequences $\mathcal{R}$}
Set $\mathcal{R} \gets \varnothing$. \\
\For{$k=1$ \KwTo $K-1$}
{
    \For{$s=1$ \KwTo $J_k$}
    {
        \uIf{$S^k(s)=0$}
        {  
           Convert matrix form $X\mathcal{W}^{Ik}_{*,s}+\mathcal{B}^{Ik}(s)>0$ into linear inequation form. \\  
           Elements in $s^{th}$ row of $\mathcal{W}^{k(k+1)}$  are set to $0$.
        }
        \Else
        {
           Convert matrix form $X\mathcal{W}^{Ik}_{*,s}+\mathcal{B}^{Ik}(s)\leq 0$ into linear inequation form. 
        }
        Add inequation to $\mathcal{R}$ as constraint.
    }
    \uIf{$k=K-1$}
    {
        Compute $\mathcal{W}^{IK} = \mathcal{W}^{I(K-1)}\mathcal{W}^{(K-1)K}$. \\
        Compute $\mathcal{B}^{IK} =  \mathcal{B}^{I(K-1)}\mathcal{W}^{(K-1)K}+\mathcal{B}^{K}$.    
    }
    \Else
    {
        Compute $\mathcal{W}^{I(k+1)} = \mathcal{W}^{Ik}\mathcal{W}^{(k(k+1)}$. \\
        Compute $\mathcal{B}^{I(k+1)} =  \mathcal{B}^{Ik}\mathcal{W}^{k(k+1)}+\mathcal{B}^{k+1}$.
    }
}
Convert matrix form $H^K = X\mathcal{W}^{IK}+\mathcal{B}^{IK}$ to linear equations. \\

\For{$j_K \in \Upsilon$}
{
    Add the following constraint to $\mathcal{R}$: \\ $w_{1(j_K)}x_1 + w_{2(j_K)}x_2 + ... + w_{M(j_K)}x_M \;\; op. \;\; (C_{j_K} - \mathcal{B}^{IK}_{j_K})$.
}

Add UDT rule's consequence to $\mathcal{R}$ as consequence.

\textbf{return} $\mathcal{R}$.
\caption{Extended C-Net Rule Extraction Algorithm (for a leaf)}\label{Alg:RuleExtraction_CNet}
\end{algorithm}

\section{Experiments}\label{section-5}

This section starts by describing the three datasets we use in the experiments. The datasets describe increasingly complex problem spaces.
The code for reproducing the results of the baseline and our proposed algorithms, as well as the codes for evaluation metrics and visualization of interpretable forms of representation extracted from the ANNs are available from the Github repository: \url{https://github.com/tudngn/EC-DT}. The tools and algorithms available there can also be used for further development of an automated explainable user interface.

\subsection{Synthetic Dataset}\label{section-5.1}
We start by constructing a feedforward network for binary classification with two input attributes. An artificial dataset is generated with polynomial relationships between the attributes. The polynomial relationship varies in complexity from one dataset to another to control the level of non-linearity in the decision boundary. A UDT may need to grow arbitrarily large before it is able to approximate a highly nonlinear function properly. We will discuss the generalization and misclassification tolerance of decision trees and our method in Section~\ref{section-6}. The classification problem is for the ANN to estimate the classes with the data distributed as

\begin{equation}
y = \begin{cases} 1, & \mbox{for } x^2_1 + x^2_2\geq 5 \\ 0, & \mbox{for } x^2_1 + x^2_2 < 5\end{cases}
\label{eq:polynomialproblem}
\end{equation}

The artificial data attributes, $x_1$ and $x_2$, are sampled uniformly within the interval of $[0,\sqrt{6}]$. Data are also sampled subject to the constraint $4 \leq x^2_1 + x^2_2 \leq 6$ so that the samples concentrate around the decision boundary. As the level of non-linearity increases, the concentration of the data increases the probability of an error due to a coarse approximation of the decision boundary.

In this problem (called the P2 problem), 10 data sets are randomly sampled each with 5000 data points. Each set is then split into training and test subsets according to the ratio 80:20. Every training subset is shuffled 10 times. Each shuffled subset is used to train an ANN. A trained model is tested with the corresponding test subset. In total, the number of trials are $10\times10=100$ models. 

The ANNs use two hidden layers with ReLU activation functions and one output unit with a sigmoid function. The number of nodes for each hidden layer is 5. The network is trained with a learning rate of 0.001, 500 epochs, and mini-batch size of 128. After a network is fully trained, decision trees using C5, Extended C-Net and EC-DT are generated to represent the network. The trees are then pruned with a pessimistic pruning algorithm identical to the one used for the C5 algorithm described in Quinlan\textquoteright s work~\cite{quinlanc5}.

\subsection{Benchmarking Datasets: UCI Data}\label{section-5.2}

We further apply our proposed methodology on benchmark datasets obtained from the UCI repository~\cite{Dua:2019}. Some properties of every data set used in this study can be found in Table~\ref{tab:datasets} (Supplementary Document, Section~\ref{section-S2}). In each problem, one ANN is built with two hidden layers, each with five hidden nodes. The number of outputs of the ANN for binary classification is 1 with sigmoid activation functions while the number for multi-class classification is the number of output classes with softmax activation functions. The training and testing schemes are similar to the process used for the P2 problem. 

\subsection{Autonomous Control Problem: Shepherding Task}\label{section-5.3}

The shepherding problem~\cite{abbass2021shepherding} is the problem of using a sheepdog to collect sheep within a paddock and herd them toward a target position. Str{\"o}mbom et al.~\cite{strombom2014solving} investigated a model that represents the interactions between the sheep and the shepherd based on simple rules. This problem motivates a form of control method for an autonomous swarm system. For example, these interactions can be used as a mode of control for unmanned aerial vehicles (UAVs) and unmanned ground vehicles (UGVs), for search-and-rescue, surveillance, and environmental management studies~\cite{abbass2021shepherding,Chen2016coordination,Minaeian2016autonomous,mathew2015planning,mathews2019supervised}.

A shepherding agent uses two main forces to characterise its movement, called collecting and driving forces. The interactions among the sheep, and between the sheep and the sheepdog are described in Section~\ref{section-S3} (Supplementary Document). The parameters of the simulation environment are described in Table~\ref{tab:EnvironmentalParameters} (Supplementary Document). For this problem, we implement a reinforcement learning model using Deep Q-Network to train two neural network models that produce the collecting (shepherding-c) and driving (shepherding-d) behaviours respectively. Each NN includes two hidden layers each with five nodes. Four inputs representing the observed states of the shepherding agent are described in detail in Section~\ref{section-S4} (Supplementary Document). Four outputs correspond to the four directions of movement (north, south, east, and west). The detailed training parameters of the reinforcement learning algorithm and the training and testing procedure are listed in Table~\ref{tab:DDQNParameters} and Section~\ref{section-S4} (Supplementary Document). For each behaviour, we train 10 NN models, from each 10 rule models are extracted with three tree extraction models. 

Later in Section~\ref{section-6.4}, we transform the set of rules representing the neural network into an interpretable visualization to improve the transparency of the mappings.

\subsection{Evaluation Metrics}\label{section-5.4}

In this paper, we focus on examining some objective metrics of interpretability which are a part of the criteria in the guidelines for designing a rule extraction system from ANNs introduced in Section~\ref{section-2-criteria}. To assess the performance of our EC-DT and Extended C-Net approaches and the baseline algorithms including classic C5 tree, LIME, s-LIME and Anchor, we introduce three types of metrics:

\begin{itemize}
    \item \textbf{Fidelity:} After extracting the rules from the trained ANNs with rule extraction algorithms, we compare the matching between outputs of the ANNs and the corresponding rule sets over test data. This metric reflects the correctness of the rule extraction models when approximating the knowledge learned by the ANN.
    For shepherding tasks, as there are no test data available, we sample random data whose dimensions are constrained within the limits defined by the size of the environment to represent possible states that an agent may encounter. The ratio of the number of matches between the outputs of a rule model and the outputs of the original neural network over the total number of samples is the \emph{fidelity} of such rule model. A fidelity of $1$ or $100\%$ means a perfect transformation from the neural network into rules.
            
    \item \textbf{Compactness}: the capability of the algorithm to represent information with the smallest model size. This could be measured by the number of rules extracted from the neural network with the rule extraction algorithms. In this paper, we also examine the average number of constraints within one rule. One set of rules may be constructed with a low number of rules. However, the final complexity of rule interpretation also depends on the number of constraints under each rule.
    
    \item \textbf{Rule interpretation}: We  investigate the decision boundaries generated by the decision trees. This could be implemented by first converting the trees into sets of rules, and second visualizing the hyper-planes corresponding to the constraints in rules on the data space.
\end{itemize}

We  analyze the computational complexity of the algorithms in Section~\ref{section-6.5}.

\section{Results and Discussion}\label{section-6}

\subsection{Performance}\label{section-6.1}

In this section, we compare the performance of different rule models with fidelity metrics, which indicates how the rule models extracted with different methods faithfully interpret the decision-making processes of the ANNs.  

\subsubsection{P2 Problem}
The fidelity of the 100 base ANNs and the corresponding rule models extracted from those networks for the P2 problem are shown in Table~\ref{tab:accuracy-5-5-UCI}. Among all rule extraction methods for ANNs, the EC-DT can maintain exactly the same performance of ANNs as expected. For the C5, LIME, s-LIME, Anchor, and Extended C-Net trees, we can observe the decrease of fidelity compared to the original performance of ANNs, by approximately 0.1\% and 1.8\% respectively. Extended C-Net preserves the performance of ANN better when compared to the baseline C5, LIME, s-LIME, and Anchor extraction method. It is important to emphasize that all transformations are deterministic, thus, variations of performance from the original ANN are not due to any stochastic variations.

\subsubsection{UCI Datasets}
The classification fidelity for UCI datasets follow a similar trend where EC-DT captures the decision-making processes of ANNs perfectly. Table~\ref{tab:accuracy-5-5-UCI} summarizes the mean and standard deviations of the fidelity of the decision trees extracted from 100 different trained ANNs on each problem. In a majority of data sets, the null hypothesis of the significant difference between the fidelity of EC-DT and the other methods is rejected with significance level of 0.05 or 0.01 (ANOVA and two-sample t-test). For the Extended C-Net algorithm, the fidelity of the generated multivariate trees are higher compared to the performance of C5 trees, LIME, s-LIME, and Anchor in seven datasets (\emph{banknote}, \emph{wdbc}, \emph{balance}, \emph{new-thyroid}, \emph{wine}, \emph{wifi}, and \emph{satimage}) by 1-5\% in average, and are equivalent to EC-DT in some cases (\emph{banknote} and \emph{wifi}). The performance of Extended C-Net is equivalent to C5, LIME, s-LIME, and Anchor on 7 other datasets (\emph{skin}, \emph{occupancy}, \emph{climate},  \emph{ionosphere}, \emph{wall-following-2}, \emph{segment}, and \emph{ecoli}). The uses of multivariate constraints between attributes of the input data can better form the necessary decision hyperplanes for classifying the output. Another interesting observation is that an increase in the number of attributes and the number of samples of datasets contributes to the deterioration of C5, LIME, s-LIME, Anchor and Extended C-Net\textquoteright s fidelity. The difference between performances of C5, LIME, s-LIME and Anchor on most data sets are not statistically significant.

\begin{table}[!htbp]
\centering
\scriptsize\addtolength{\tabcolsep}{-3pt}
\begin{threeparttable}
\caption{Accuracy of ANN models and fidelity of decision rules generated from C5, LIME, s-LIME, Anchor Extended C-Net and EC-DT algorithms.}
\label{tab:accuracy-5-5-UCI}
\begin{tabular}{lccccccc}
\begin{tabular}[c]{@{}c@{}}\textbf{Data set}\end{tabular} &
\begin{tabular}[c]{@{}c@{}}\textbf{ANN} \\ \textbf{Accuracy} \\ ($\mu \pm \sigma \%$)\end{tabular} &
\multicolumn{6}{c}{\textbf{Fidelity of Rule Extraction Algorithm for NN}} \\ \cline{3-8}
  & &
\begin{tabular}[c]{@{}c@{}}\textbf{C5} \\ ($\mu \pm \sigma \%$)\end{tabular}
& 
\begin{tabular}[c]{@{}c@{}}\textbf{LIME} \\ ($\mu \pm \sigma \%$)\end{tabular}
&
\begin{tabular}[c]{@{}c@{}}\textbf{s-LIME} \\ ($\mu \pm \sigma \%$)\end{tabular}
&
\begin{tabular}[c]{@{}c@{}}\textbf{Anchor} \\ ($\mu \pm \sigma \%$)\end{tabular}
&
\begin{tabular}[c]{@{}c@{}}\textbf{Extended C-Net} \\ ($\mu \pm \sigma \%$)\end{tabular} 
& 
\begin{tabular}[c]{@{}c@{}}\textbf{EC-DT} \\ ($\mu \pm \sigma \%$)\end{tabular} \\
\hline
P2
                            & 93.81$\pm$5.45
                            
                            & 97.47$\pm$9.22
                            
                            & 89.10$\pm$7.36
                            
                            & 90.23$\pm$7.47

                            & 96.85$\pm$7.16

                            & 98.63$\pm$10.12                                      
                            & 100.00$\pm$0.00 
                            
                            \\

skin                        
                            & 94.43$\pm$8.93
    
                            & 91.86$\pm$11.18 

                            & 84.33$\pm$3.57
                            
                            & 83.40$\pm$2.32

                            & 87.85$\pm$9.27
                                                      
                            & 92.62$\pm$14.83                                                       
                            & \textbf{100.00$\pm$0.00$^\dag$}                                              
                            \\
banknote   
                            & 99.84$\pm$0.49
    
                            & 98.88$\pm$0.96                                  
                            
                            & 98.78$\pm$0.72
                            
                            & 98.78$\pm$0.72
                            
                            & 97.35$\pm$0.71
                                                        
                            & 99.80$\pm$0.48                                                  
                            & 100.00$\pm$0.00                                          
                            \\  
occupancy 
                            & 98.86$\pm$0.15
    
                            & 99.95$\pm$0.08 
                            
                            & 83.67$\pm$3.90
                            
                            & 85.47$\pm$3.02

                            & 88.85$\pm$0.15
                                                        
                            & 99.96$\pm$0.07                                                          
                            & 100.00$\pm$0.00                                                          
                            \\
wdbc   
                            & 94.57$\pm$1.88
    
                            & 96.49$\pm$2.51    
                            
                            & 96.19$\pm$1.09
                            
                            & 96.20$\pm$1.09
                            
                            & 92.54$\pm$1.50
                                                        
                            & 96.67$\pm$1.94                                                          
                            & \textbf{100.00$\pm$0.00$^*$}                                              
                            \\
climate  
                            & 91.01$\pm$3.02
    
                            & 94.04$\pm$7.98                                                              
                            & 90.12$\pm$2.43
                            
                            & 90.12$\pm$2.43
                         
                            & 84.33$\pm$8.36
                                                         
                            & 94.17$\pm$8.67                                                          
                            & \textbf{100.00$\pm$0.00$^\dag$}                                           
                            \\
ionosphere 
                            & 93.11$\pm$3.38
    
                            & 95.86$\pm$3.03                                                              
                            & 83.33$\pm$2.43
                            
                            & 83.33$\pm$2.43
                          
                            & 90.65$\pm$3.17
                                                        
                            & 93.29$\pm$5.88                                                          
                            & \textbf{100.00$\pm$0.00$^*$}                                           
                            \\
balance 
                            & 95.12$\pm$3.22
    
                            & 84.64$\pm$5.17                                                              
                            & 69.07$\pm$3.29
                            
                            & 69.07$\pm$3.29
                          
                            & 78.72$\pm$3.32
                                                        
                            & 97.52$\pm$2.48                                                          
                            & \textbf{100.00$\pm$0.00$^*$}                                               
                            \\
new-thyroid 
                            & 97.12$\pm$2.98
    
                            & 94.44$\pm$2.64                                                              
                            & 93.02$\pm$3.80
                            
                            & 93.02$\pm$3.80
                          
                            & 90.93$\pm$8.91
                                                        
                            & 96.53$\pm$5.14                                                          
                            & \textbf{100.00$\pm$0.00$^\dag$}                                           
                            \\
wine  
                            & 97.13$\pm$2.29
    
                            & 94.50$\pm$4.46                                                              
                            & 88.70$\pm$6.55
                            
                            & 88.70$\pm$6.55
                          
                            & 91.94$\pm$7.19
                                                        
                            & 98.17$\pm$3.60                                                          
                            & \textbf{100.00$\pm$0.00$^\dag$}                                              
                            \\  
                            
wall-following-2
                            & 97.82$\pm$2.84
    
                            & 97.84$\pm$3.10                                                              
                            & 83.80$\pm$1.49
                            
                            & 81.17$\pm$2.29
                          
                            & 84.58$\pm$1.26
                                                        
                            & 98.40$\pm$3.70                                                          
                            & \textbf{100.00$\pm$0.00$^\dag$}                                          
                            \\
wall-following-4 
                            & 97.59$\pm$1.90 
    
                            & 99.63$\pm$0.69                                                              
                            & 81.60$\pm$2.12
                            
                            & 82.67$\pm$2.53
                          
                            & 87.08$\pm$2.97
                                                
                            & 99.57$\pm$1.84                                                          
                            & 100.00$\pm$0.00                                                  
                            \\

wifi  
                            & 97.61$\pm$1.17
    
                            & 98.75$\pm$0.64                                                              
                            & 95.50$\pm$0.50
                            
                            & 95.62$\pm$0.41
                           
                            & 96.25$\pm$0.47
                                                       
                            & 99.37$\pm$0.81                                                  
                            & 100.00$\pm$0.00                                                  
                            \\
dermatology 
                            & 95.34$\pm$3.70
    
                            & 99.39$\pm$0.98                                                              
                            & 84.25$\pm$2.96
                            
                            & 84.25$\pm$2.96
                
                            & 93.73$\pm$3.06
                                                       
                            & 98.51$\pm$3.31                                                          
                            & 100.00$\pm$0.00                                               
                            \\                            

satimage   
                            & 89.21$\pm$1.01
    
                            & 96.83$\pm$1.93                                                              
                            & 81.27$\pm$3.81
                            
                            & 82.28$\pm$4.73
                          
                            & 80.56$\pm$6.21
                                                        
                            & 98.86$\pm$0.90                                                         
                            & \textbf{100.00$\pm$0.00$^*$}                                           
                            \\
                            
segment   
                            & 95.44$\pm$1.35
    
                            & 99.79$\pm$1.26                                                              
                            & 84.79$\pm$2.17
                            
                            & 84.79$\pm$2.17
                          
                            & 85.17$\pm$1.77
                                                        
                            & 98.96$\pm$1.58                                                          
                            & 100.00$\pm$0.00                                              
                            \\

ecoli    
                            & 89.66$\pm$4.85
    
                            & 94.22$\pm$2.09                                                              
                            & 86.07$\pm$3.12
                            
                            & 86.07$\pm$3.12
                          
                            & 79.61$\pm$0.94
                                                        
                            & 93.13$\pm$4.25                                                          
                            & \textbf{100.00$\pm$0.00$^*$}                                               
                            \\
                            
HIGGS
                            & 72.08$\pm$0.46
    
                            & 84.89$\pm$2.83
                                                        
                            & 71.45$\pm$5.30
                            
                            & 72.22$\pm$5.71

                            & 78.83$\pm$5.14
                            
                            & 86.84$\pm$3.22
                            
                            & \textbf{100.00$\pm$0.00$^*$}
                            
                            \\
                            
shepherding-c 
                            & N/A
    
                            & 94.16$\pm$0.16  
                                                        
                            & 95.35$\pm$0.15
                            
                            & 95.22$\pm$0.26

                            & 94.74$\pm$0.08
                            
                            & 99.45$\pm$0.07                                                        
                            & \textbf{100.00$\pm$0.00$^*$}                                              
                            \\
shepherding-d  
                            & N/A
    
                            & 92.67$\pm$0.12                                
                                                        
                            & 94.27$\pm$0.27
                            
                            & 94.14$\pm$0.19

                            & 93.51$\pm$0.16
                                                      
                            & 98.34$\pm$0.10                                                  
                            & \textbf{100.00$\pm$0.00$^*$}                                           
                            \\  

\end{tabular}
\begin{tablenotes}
      \small
      \item Figures in \textbf{bold} are the best among methods.\\
      $^{\dag}$ significantly better than its counterparts at significant level of 0.05.\\
      $^*$ significantly better than its counterparts at significant level of 0.01.
    \end{tablenotes}
\end{threeparttable}
\end{table}

\subsubsection{Shepherding Problem}
Regarding shepherding problems, the mean and standard deviations of the fidelity of 100 decision trees (for each method) extracted from 10 trained ANNs for collecting behaviour and 10 trained ANNs for driving behaviour are also reported in Table~\ref{tab:accuracy-5-5-UCI}. The results suggest that the EC-DT algorithm can generate rule models with $100\%$ fidelity. Therefore the EC-DT tree models can reproduce perfectly the performance of the ANNs for both shepherding sub-tasks. Fidelities of Extended C-Net and C5 trees are roughly $1.5\%$, $6.5\%$ and $7.5\%$ lower than that of EC-DT trees on average.

\subsection{Compactness and Fidelity}\label{section-6.2}

The analysis of the dimensions of the set of rules extracted from the ANNs provides us the information on the compactness achieved by the proposed methods compared to the baseline rule extraction algorithms. There are two main indicators: the number of extracted rules, and the number of constraints/premises in each rule. The former indicator indicates the size of the rule models. Depending on the system, one can select a more comprehensive model with large size to achieve better fidelity or select a more compact model if the system does not have sufficient memory. As there is a strong connection between the model size and the fidelity of the model, we analyze both properties of the sets of rules extracted by different methods. On the other hand, the number of constraints/premises in each rule directly relates to the interpretation for a given input-output pair of the original ANNs, therefore influences the interpretability of the rule models. A complex set of constraints might provide a faithful interpretation of the actual decision-making processes from the original model. However, a higher number of constraints in each rule can make its interpretation more complex, and hence reduces the transparency of the interpretation.

In this section, we mainly discuss the connections between two compactness indicators and the fidelity of the rule models. The relationships between the individual interpretation and transparency will be discussed in Section~\ref{section-6.4}.

\subsubsection{Number of rules}

It can be demonstrated from the data in Table~\ref{tab:tree-size-UCI} that the number of rules extracted by Extended C-Net are significantly lower than the number of rules extracted by C5, LIME, s-LIME, and Anchor algorithms in most problems while maintaining an equivalent or higher fidelity in most problems. The number of leaves in C5, and the number of rules in LIME, s-LIME, and Anchor algorithms are 3-10 times higher than the Extended C-Net, which implies that a much greater number of decision hyperplanes is used for the classification problem. The differences between Extended C-Net tree sizes and C5, LIME, s-LIME, and Anchor's size of sets of rules, according to the figures, varies dramatically from around 1 (e.g.\emph{occupancy} and \emph{wine}) up to 40 (e.g. \emph{skin}) depending on the complexity and nonlinearity of the problem. In cases with such large differences, the fidelity of the Extended C-Net are also equivalent or better than the fidelity of C5, LIME, s-LIME, and Anchor. This is due to the capability of Extended C-Net to generate multivariate trees for representing the behaviour of the ANNs better, much like the EC-DT. The numbers of rules generated from LIME, s-LIME, and Anchor algorithm are surprisingly higher than any other algorithm. The reason for this phenomenon might be because LIME, s-LIME, and Anchor creates a very high number of rules each of which only covers a small area in the problem space. Also, the local rules extracted by LIME, s-LIME, and Anchor overlap with one another, which results in a very high number of rules in total. According to the ``stability'' criteria reported by Carvalho et al.~\cite{carvalho2019machine}, although these rules might be correct in majority of cases, the model is not stable as similar interpretation is expected for similar input instances.

For \emph{shepherding-c} and \emph{shepherding-d} problems, C5 trees, however, need an average number of leaves of 3 to 4 times larger than those of EC-DT or Extended C-Net models. This is due to the fact that for autonomous control problems like shepherding, the representative state space can be much more complicated than the problem space of some conventional classification problems. For shepherding problem, the state space can be segmented into a huge number of smaller sub-spaces based on distance or direction between the shepherd and the sheep. Each of those sub-space might include all possible output classes whose distributions overlapping with each other. Therefore, an algorithm that creates axis-parallel decision polytopes like univariate C-5 is not appropriate to model such complex data distribution.

\begin{table}[!htbp]
\centering
\scriptsize\addtolength{\tabcolsep}{-3pt}
\begin{threeparttable}
\caption{Means and standard deviations of number of rules extracted with C5, LIME, s-LIME, Anchor, Extended C-Net, and EC-DT for different problems.}
\label{tab:tree-size-UCI}
\begin{tabular}{lcccccc}
\textbf{Data set} & \textbf{C5} & \textbf{LIME} & \textbf{s-LIME} & \textbf{Anchor} & \textbf{Extended C-Net} & \textbf{EC-DT} \\ \hline

P2                      & 38.73$\pm$11.87                                                             
                        & 75.53$\pm$8.12
                        
                        & 74.72$\pm$9.25
                        
                        & 51.13$\pm$9.36

                        & \textbf{11.12$\pm$6.00$^{\dag}$}                    
                        
                        & 13.40$\pm$6.35                                      
                        \\

skin                    
                        & 48.25$\pm$30.77                                   
                        
                        & 89.52$\pm$17.12
                        
                        & 86.75$\pm$16.13
                
                        & 82.30$\pm$15.45
           
                        & \textbf{7.65$\pm$8.49$^*$}                                               
                        & 28.93$\pm$20.48                                                     
                        \\
banknote
                        & 12.07$\pm$2.36
                        
                        & 115.67$\pm$2.49
                        
                        & 115.67$\pm$2.49
                                  
                        & 43.41$\pm$6.18
                  
                        & \textbf{3.91$\pm$1.33$^*$}                                                                               
                        & 45.72$\pm$9.71                                                           
                        \\
occupancy
                        & 2.52$\pm$1.35                                       
                        
                        & 179.67$\pm$6.43
                        
                        & 171.67$\pm$5.19
                
                        & 35.18$\pm$6.35
                   
                        & \textbf{2.05$\pm$0.41$^*$}                                         
                        & 13.50$\pm$7.36                                                        
                        \\
wdbc
                        & 8.90$\pm$1.58                                       
                        
                        & 111.31$\pm$2.45
                        
                        & 111.28$\pm$2.45
                
                        & 46.20$\pm$7.16
                 
                        & \textbf{3.08$\pm$1.15$^*$}
                   
                        & 11.39$\pm$4.52                                                           
                        \\
climate
                        & 6.92$\pm$3.19                                       
                        
                        & 108.00$\pm$0.00
                        
                        & 108.00$\pm$0.00
                
                        & 48.56$\pm$8.93
               
                        & \textbf{2.87$\pm$1.35$^*$}
                   
                        & 16.96$\pm$11.85                                                      
                        \\
ionosphere
                        & 7.59$\pm$1.58                                       
                        
                        & 70.00$\pm$0.00
                        
                        & 70.00$\pm$0.00
                
                        & 58.13$\pm$11.56
                 
                        & \textbf{2.92$\pm$1.28$^*$}
                 
                        & 28.01$\pm$10.16                                                     
                        \\
balance
                        & 18.98$\pm$4.37                                      
                        
                        & 101.96$\pm$2.62
                        
                        & 101.96$\pm$2.62
                
                        & 76.82$\pm$14.61
                  
                        & \textbf{9.19$\pm$4.06$^*$}
                
                        & 20.39$\pm$11.09                                                     
                        \\
                        
new-thyroid
                        & 4.73$\pm$0.97                                       
                        
                        & 36.35$\pm$0.94
                        
                        & 36.35$\pm$0.94
                
                        & 35.87$\pm$7.14
                
                        & \textbf{2.97$\pm$0.36$^*$}
               
                        & 9.53$\pm$3.53                                                       
                        \\ 
wine
                        & 4.51$\pm$0.50                                       
                        
                        & 36.00$\pm$0.00
                        
                        & 36.00$\pm$0.00
                
                        & 56.50$\pm$8.19
             
                        & \textbf{3.00$\pm$0.00$^*$}
              
                        & 27.39$\pm$4.84                                                       
                        \\                        
wall-following-2
                        & \textbf{9.92$\pm$3.99$^*$}                          
                        
                        & 52.33$\pm$7.11
                        
                        & 57.67$\pm$5.81
                
                        & 37.12$\pm$2.98
                  
                        & 18.87$\pm$9.20
                   
                        & 12.15$\pm$4.61                                      
                        
                        \\
wall-following-4
                        & \textbf{11.38$\pm$4.23$^*$}                         
                        
                        & 185.19$\pm$7.85
                        
                        & 195.41$\pm$4.78
                
                        & 79.82$\pm$10.51
                 
                        & 21.71$\pm$10.39
                
                        & 16.50$\pm$6.49                                                     
                        \\
wifi
                        & 13.96$\pm$3.22                                      
                        
                        & 316.47$\pm$28.54
                        
                        & 316.83$\pm$26.98
                
                        & 89.50$\pm$14.81
                  
                        & \textbf{8.09$\pm$1.83$^*$}
                   
                        & 14.50$\pm$5.56                                                         
                        \\
dermatology
                        & 8.06$\pm$0.87                                       
                        
                        & 73.52$\pm$0.10
                        
                        & 73.52$\pm$0.10
                
                        & 47.41$\pm$5.54

                        & \textbf{6.29$\pm$0.55$^*$}
                   
                        & 27.96$\pm$8.19                                                          
                        
                        \\ 
satimage
                        & 70.83$\pm$6.76                                      
                        
                        & 468.30$\pm$60.81
                        
                        & 465.42$\pm$59.41
                
                        & 374.17$\pm$52.23
                
                        & \textbf{38.62$\pm$6.93$^*$}
                
                        & 80.86$\pm$18.63                                                          
                        
                        \\ 
segment
                        & 24.20$\pm$3.42                                      
                        
                        & 89.10$\pm$7.12
                        
                        & 86.65$\pm$6.13
                
                        & 159.50$\pm$23.71
                                                                    
                        & 23.60$\pm$4.47
               
                        & 37.85$\pm$11.71                                                     
                        \\ 
ecoli
                        & 10.67$\pm$2.09                                      
                        
                        & 59.67$\pm$2.08
                        
                        & 59.67$\pm$2.08
                
                        & 67.81$\pm$8.63
                  
                        & \textbf{8.88$\pm$1.96$^*$}
                  
                        & 13.84$\pm$6.48                                      
                        \\ 
                        
HIGGS
                        & 86.42$\pm$5.69
                        
                        & 415.52$\pm$71.43
                        
                        & 427.73$\pm$69.36
                
                        & 387.29$\pm$63.16
                        
                        & \textbf{43.10$\pm$8.25$^*$}
                        
                        & 109.90$\pm$0.00
                        
                        \\

shepherding-c
                        & 791.30$\pm$26.04                                    
                        
                        & 2560.21$\pm$61.33
                        
                        & 2498.46$\pm$72.55
                
                        & 2488.35$\pm$65.12
                  
                        & \textbf{136.50$\pm$9.56$^*$}
                    
                        & 232.00$\pm$28.40                                                      
                        \\ 
shepherding-d
                        & 1098.00$\pm$41.46                                   
                        
                        & 3281.69$\pm$101.57
                        
                        & 3254.90$\pm$96.30
                
                        & 3175.25$\pm$81.29
             
                        & 347.00$\pm$18.26
                
                        & \textbf{282.00$\pm$51.28$^*$}                                        
                        
\end{tabular}
\begin{tablenotes}
      \small
      \item Figures in \textbf{bold} are the best among methods.\\
      $^{\dag}$ significantly better than its counterparts at significant level of 0.05.\\
      $^*$ significantly better its counterparts at significant level of 0.01.
    \end{tablenotes}
\end{threeparttable}
\end{table}

\subsubsection{Number of constraints in each rule}
 
As illustrated in Table~\ref{tab:constraints-size-UCI}, the constraints that one can find on average at each rule of the C5 trees is much lower than the others. With less than 2-5 constraints per rule, the difference between the number of constraints per rule in C5, LIME, s-LIME, Anchor and Extended C-Net is significant (ANOVA and two-sample t-test in both one and two tails with significant level of 0.01). The number of constraints in each rule in LIME, s-LIME, and Anchor on average are higher than those of C5. The reason for the significantly higher number of constraints per rule of Extended C-Net and EC-DT comes from more comprehensive representation of the input-output constraints to maintain a higher fidelity than other methods.

The low number of constraints per rule reflects the simplicity of using only some of the most relevant attributes for rule extraction. The use of less attributes in C5 and Anchor cannot achieve an equivalent level of fidelity, compared to Extended C-Net or EC-DT, given the nature of C5, LIME, s-LIME, and Anchor univariate rules which produces axis parallel decision hyperplanes in the problem space. However, for some problems such as \emph{occupancy}, \emph{wall-following-2}, \emph{wall-following-4}, \emph{dermatology}, \emph{segment}, and \emph{ecoli}, it is noticeable that even with a much lower number of constraints per rule, the C5 and Anchor algorithm still shows an equivalent or higher fidelity compared to Extended C-Net (though both are not comparable to EC-DT in terms of fidelity). This phenomenon raises an important question: in which situation, the use of axis parallel decision hyperplanes induced from simple univariate rules extracted by Anchor can be applied without sacrificing much fidelity? To answer this question, we investigate the decision boundary complexity in Section~\ref{section-6.3}.

\begin{table}[!htbp]
\centering
\footnotesize
\begin{threeparttable}
\caption{Means and standard deviations of number of constraints under one rule extracted with C5, LIME, s-LIME, Anchor, Extended C-Net, and EC-DT algorithms.}
\label{tab:constraints-size-UCI}
\begin{tabular}{lcccccc}
\textbf{Data set} & \textbf{C5} & \textbf{LIME} & \textbf{s-LIME} & \textbf{Anchor} & \textbf{Extended C-Net} & \textbf{EC-DT} \\ \hline                                                      

P2                      
                        & \textbf{1.91$\pm$0.16$^*$}                        
                        
                        & 3.46$\pm$0.05
                        
                        & 3.37$\pm$0.04
                        
                        & 2.18$\pm$0.06
                      
                        & 6.69$\pm$1.05                                                             
                        & 11.00$\pm$0.01 
                        
                        \\

skin                    
                      
                        & \textbf{2.59$\pm$0.98$^*$}  
                        
                        & 4.64$\pm$0.82
                        
                        & 4.60$\pm$0.88
                                        
                        & 3.40$\pm$0.13
                          
                        & 5.30$\pm$2.59                                                      
                        & 11.00$\pm$0.02                                                             
                        
                        \\
banknote
                     
                        & \textbf{2.33$\pm$0.10$^*$}                          
                        
                        & 6.20$\pm$1.05
                        
                        & 6.20$\pm$1.05
                        
                        & 2.68$\pm$0.17
                          
                        & 6.46$\pm$0.27                                                     
                        & 10.99$\pm$0.01                                                             
                        \\
occupancy
                     
                        & \textbf{1.18$\pm$0.41$^*$}                          
                        
                        & 7.34$\pm$1.10
                        
                        & 7.39$\pm$1.12
                                       
                        & 2.11$\pm$0.06
                
                        & 6.01$\pm$0.09                                                             
                        & 11.00$\pm$0.02                                                            
                        
                        \\
wdbc
                        & \textbf{2.11$\pm$0.25$^*$}   
                        
                        & 15.17$\pm$2.50
                        
                        & 15.17$\pm$2.50
                        
                        & 4.43$\pm$0.15

                        & 6.28$\pm$0.28                                                         
                        & 11.00$\pm$0.01                                                            
                        \\
climate
                        & \textbf{1.95$\pm$0.76$^*$}                          
                        
                        & 22.48$\pm$1.88
                        
                        & 22.48$\pm$1.88
                                        
                        & 3.64$\pm$0.10

                        & 5.46$\pm$1.97                                                              
                        & 11.00$\pm$0.01                                                            
                        \\
ionosphere
                        & \textbf{1.65$\pm$0.48$^*$}                          
                        
                        & 47.61$\pm$8.02
                        
                        & 47.61$\pm$8.02
                                       
                        & 2.91$\pm$0.06

                        & 6.13$\pm$0.79                                                              
                        & 11.00$\pm$0.01                                                            
                        
                        \\
balance
                     
                        & \textbf{2.85$\pm$0.27$^*$}                        
                        
                        & 5.72$\pm$0.98
                        
                        & 5.72$\pm$0.98
                        
                        & 3.48$\pm$0.18

                        & 7.26$\pm$0.41                                                              
                        & 10.00$\pm$0.01                                                         
                        \\
                        
new-thyroid
                    
                        & \textbf{1.55$\pm$0.36$^*$}                        
                        
                        & 7.65$\pm$1.32
                        
                        & 7.65$\pm$1.32
                        
                        & 3.50$\pm$0.10

                        & 6.27$\pm$0.77                                                            
                        & 10.00$\pm$0.02                                                            
                        
                        \\ 
wine
                    
                        & \textbf{1.72$\pm$0.22$^*$}                          
                        
                        & 19.13$\pm$1.80
                        
                        & 19.13$\pm$1.80
                                     
                        & 2.69$\pm$0.13

                        & 6.48$\pm$0.17                                                             
                        & 9.99$\pm$0.02                                                            
                        
                        \\                        
wall-following-2
                         
                        & \textbf{2.08$\pm$0.56$^*$}                        
                        
                        & 2.89$\pm$0.67
                        
                        & 2.97$\pm$0.63
                        
                        & 2.30$\pm$0.06

                        & 7.23$\pm$0.98                                                                                            
                        & 9.00$\pm$0.02                                                            
                        
                        \\
wall-following-4
                          
                        & \textbf{2.35$\pm$0.45$^*$}                        
                        
                        & 6.10$\pm$1.09
                        
                        & 6.09$\pm$1.05
                        
                        & 4.14$\pm$0.07

                        & 7.63$\pm$0.45                                            
                        & 9.00$\pm$0.01                                                            
                        
                        \\
wifi
      
                        & \textbf{2.83$\pm$0.30$^*$}                          
                        
                        & 10.77$\pm$1.71
                        
                        & 10.77$\pm$1.71
                             
                        & 4.51$\pm$0.27

                        & 7.15$\pm$0.24                                                          
                        & 9.00$\pm$0.01                                                            
                        
                        \\
dermatology
                    
                        & \textbf{2.68$\pm$0.20$^*$}                          
                        
                        & 37.00$\pm$2.11
                        
                        & 37.00$\pm$2.11
                              
                        & 3.21$\pm$0.09

                        & 7.36$\pm$0.81                                          
                        & 8.99$\pm$0.02                                                           
                        
                        \\ 
satimage
                     
                        & \textbf{4.96$\pm$0.21$^*$}                          
                        
                        & 55.06$\pm$8.62
                        
                        & 54.92$\pm$8.60
                                  
                        & 7.25$\pm$0.16

                        & 8.53$\pm$0.21                                                   
                        & 8.99$\pm$0.01                                                            
                        
                        \\ 
segment
             
                        & \textbf{3.46$\pm$0.33$^*$}                          
                        
                        & 27.21$\pm$3.69
                        
                        & 27.15$\pm$3.67
                        
                        & 4.10$\pm$0.05

                        & 8.29$\pm$0.30                                                               
                        & 8.99$\pm$0.01                                                            
                        
                        \\ 
ecoli
                     
                        & \textbf{2.80$\pm$0.33$^*$}                          
                        
                        & 7.46$\pm$1.08 
                        
                        & 7.46$\pm$1.08
                          
                        & 4.91$\pm$0.12

                        & 7.40$\pm$0.30                                                        
                        & 9.00$\pm$0.01                                     

                        \\ 
                        
HIGGS
                     
                        & \textbf{6.73$\pm$0.26$^*$}                          
                        
                        & 19.45$\pm$2.31
                        
                        & 20.67$\pm$3.18
                                  
                        & 7.42$\pm$0.35

                        & 9.33$\pm$0.27                                                   
                        & 11.00$\pm$0.01                                                            
                        
                        \\ 
shepherding-c
                       
                        & 4.98$\pm$0.15                                       
                        
                        & 7.72$\pm$0.83   
                        
                        & 7.98$\pm$0.91
                                   
                        & 4.59$\pm$0.15

                        & \textbf{3.69$\pm$0.09$^*$}                             
                        & 9.00$\pm$0.01
                        
                        \\ 
shepherding-d
                        
                        & 5.31$\pm$0.17                                       
                        
                        & 9.15$\pm$1.40
                        
                        & 9.81$\pm$1.15
                        
                        & 5.42$\pm$0.18

                        & \textbf{4.06$\pm$0.11$^*$}                                                      
                        & 9.00$\pm$0.01

\end{tabular}
\begin{tablenotes}
      \small
      \item Figures in \textbf{bold} are the best among methods.\\
      $^{\dag}$ significantly better than its counterparts at significant level of 0.05.\\
      $^*$ significantly better its counterparts at significant level of 0.01.
    \end{tablenotes}
\end{threeparttable}
\end{table}

\subsection{Decision Boundary Complexity}\label{section-6.3}

While Extended C-Net provides the most compact trees and lower size of rule sets, C5, LIME, s-LIME, and Anchor provide more interpretability with much lower number of constraints for each rule. That raises a question of when to use a classic such as C5 or a local approximation algorithm like Anchor instead of a more complex and comprehensible model like Extended C-Net or EC-DT. In other words, how to decide the best compromise between simplicity versus fidelity. The analysis of the decision boundary sheds some light on this issue. 

\begin{figure}[!ht]
\centering
    \subfloat[Occupancy]{\includegraphics[width=0.5\columnwidth]{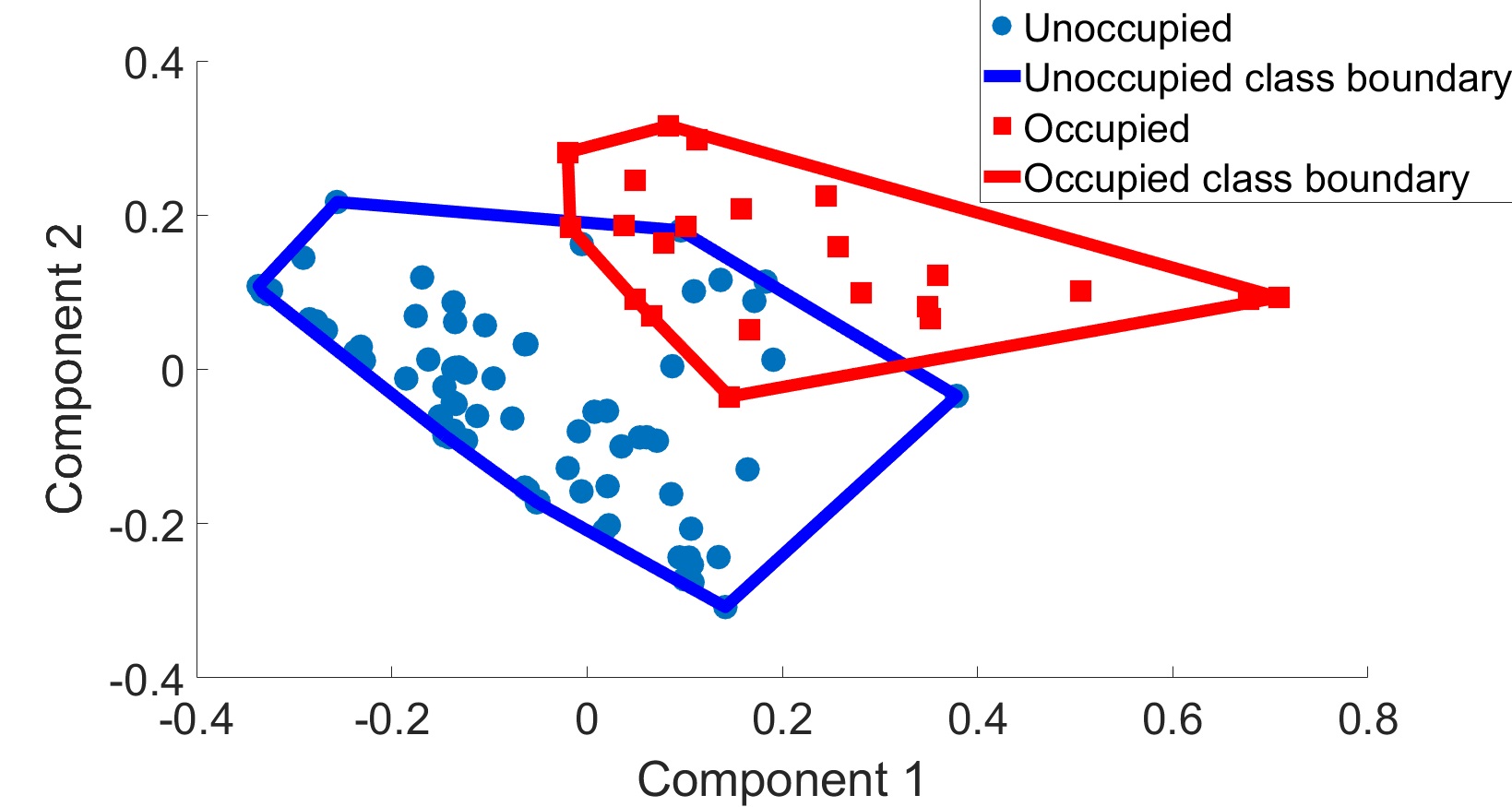}}
    \subfloat[Skin]{\includegraphics[width=0.5\columnwidth]{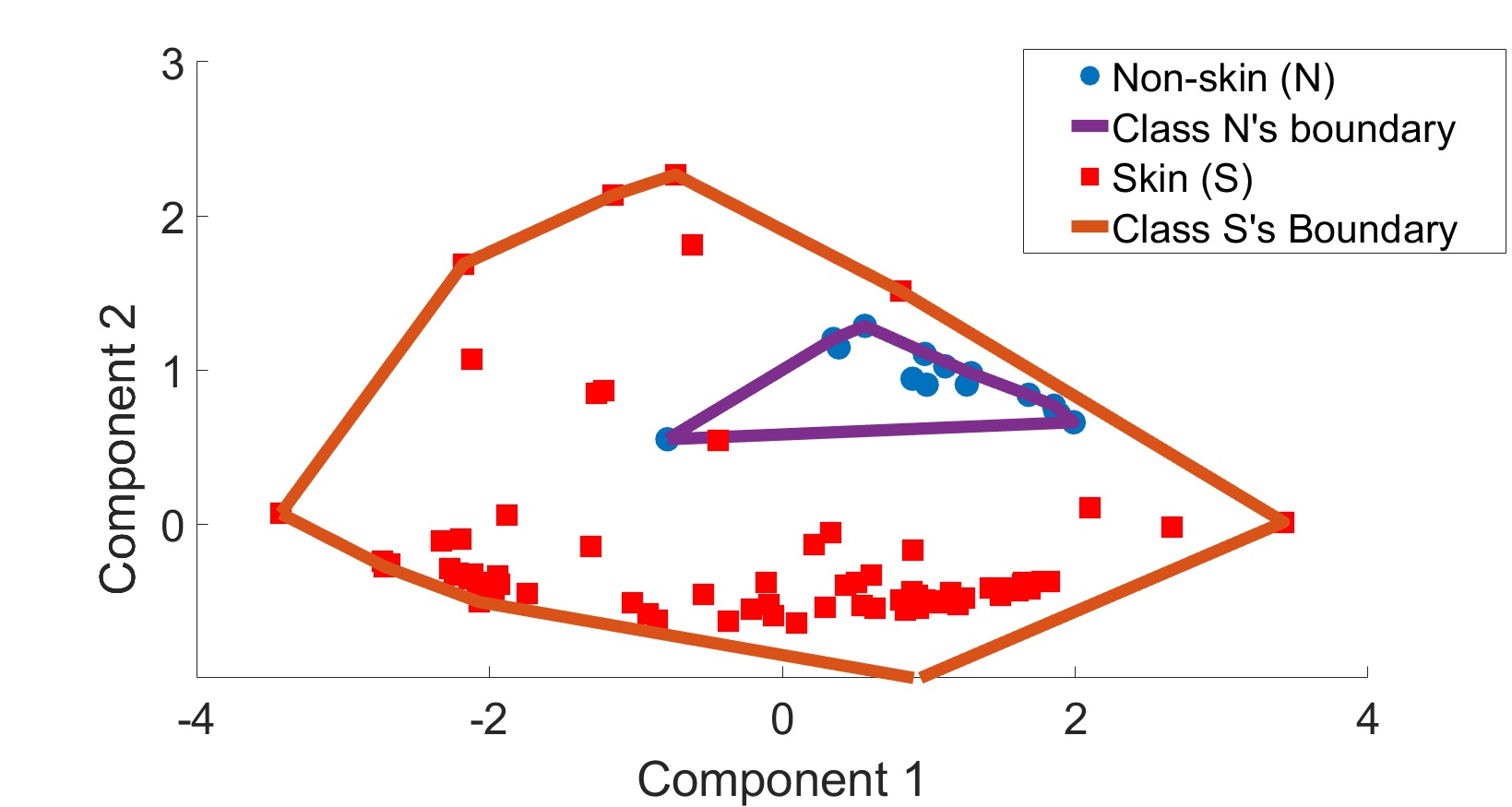}}

    \subfloat[Wall following (4 inputs)]{\includegraphics[width=0.5\columnwidth]{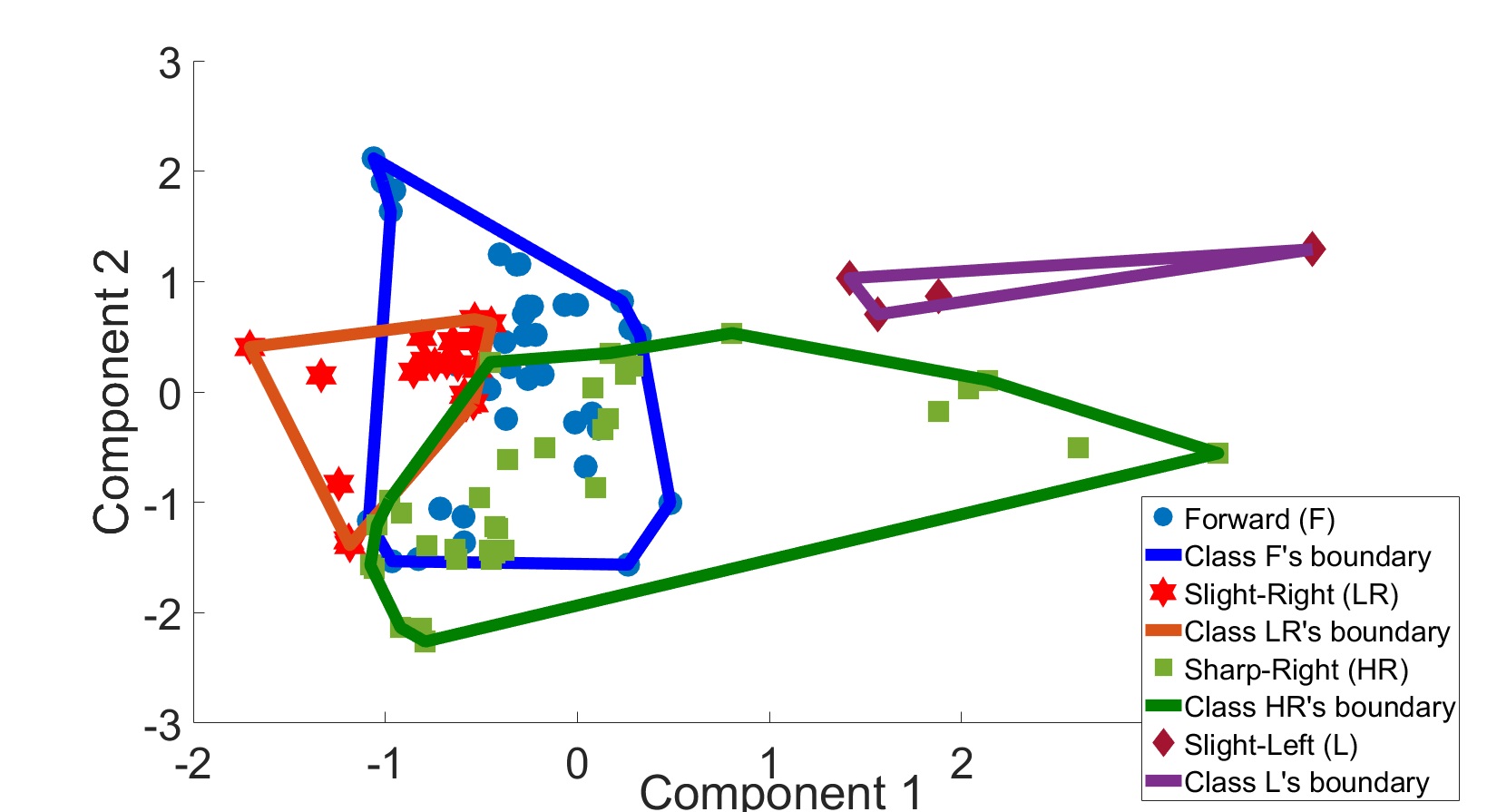}}
    \subfloat[Balance]{\includegraphics[width=0.5\columnwidth]{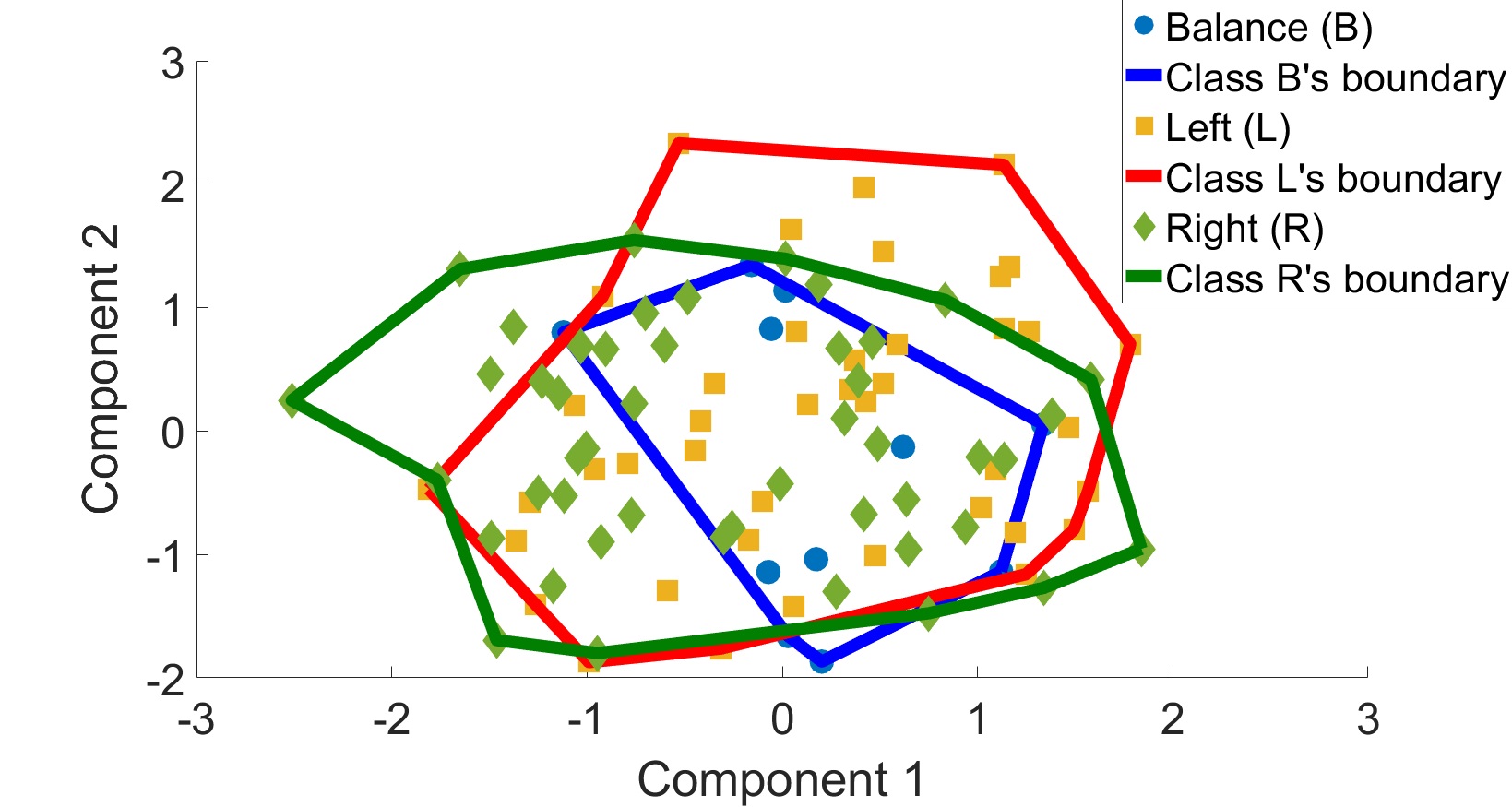}}
    \caption{Two PCA components with highest variances of different datasets.}
\label{fig:data_distribution}
\end{figure}

For analyzing the complexity of the data spaces and class distributions, we apply a Principle Component Analysis (PCA) transformation on each data set and visualize the two largest components with the largest variance. Figure~\ref{fig:data_distribution} illustrates the distribution of classes according to two chosen principle components in two binary classification problem (\emph{occupancy} and \emph{skin}) and classification problem with more than two target labels (\emph{wall-following-4} and \emph{balance}). It can be seen that the data classes in the problem of \emph{occupancy} and \emph{wall-following-4} are more linearly separable than the other problems where the class coverage strongly overlaps with one another. Due to this linear-separability, it is feasible to use a simple tree with low number of leaves and constraints under each rule to represent the classification rules and hence achieve a higher fidelity with the performance of ANNs. The visualization of PCA components of different classes in remaining data sets can be found in Section~\ref{section-S6} (Supplementary document). 

According to the figures in Table~\ref{tab:accuracy-5-5-UCI} and Table~\ref{tab:tree-size-UCI}, the fidelity of linearly-separable problems such as \emph{occupancy} and \emph{wall-following-4} are not significantly different among all methods, while the number of constraints under each leaf/rule on average of C5 and Anchor is equivalent or much lower than the ones of Extended C-Net. In these cases, the use of C5 or Anchor for extracting rules from the ANNs is more appropriate as they achieve acceptable performance with better simplicity. On the other hand, for more complex problems with non-linear separable properties, it is less accurate when using C5 and Anchor to extract the rules. In the \emph{skin} problem, to achieve around 86\% fidelity, the C5 trees have to use up to nearly 50 leaves each with more than 2.5 rules on average. As another example, the C5 generates approximately 19 rules, each includes around 3 constraints, to only achieve less than 80\% fidelity on average in \emph{balance} problem. For the same problem, the Extended C-Net reaches more than 92\% fidelity with half the number of rules of C5, but with more than 7 constraints in each rule. Therefore, it is favorable to use C5 with simple rule sets for linearly-separable datasets with low dimensions while using Extended C-Net or EC-DT if one favors higher fidelity and understanding of correlation among a large number of attributes of input space. The number of rules generated when using LIME, s-LIME, and Anchor algorithm is much larger than EC-DT and Extedned C-Net. However, as LIME, s-LIME, and Anchor does not need to build an entire set of rules that covers the entire problem space, it can be use with less complex or linearly-separable datasets, which would enhance the explanation speed and the interpretability while maintaining a relatively high fidelity.

\begin{figure}[!ht]
\centering
\subfloat[C5]{%
  \includegraphics[width=0.5\columnwidth]{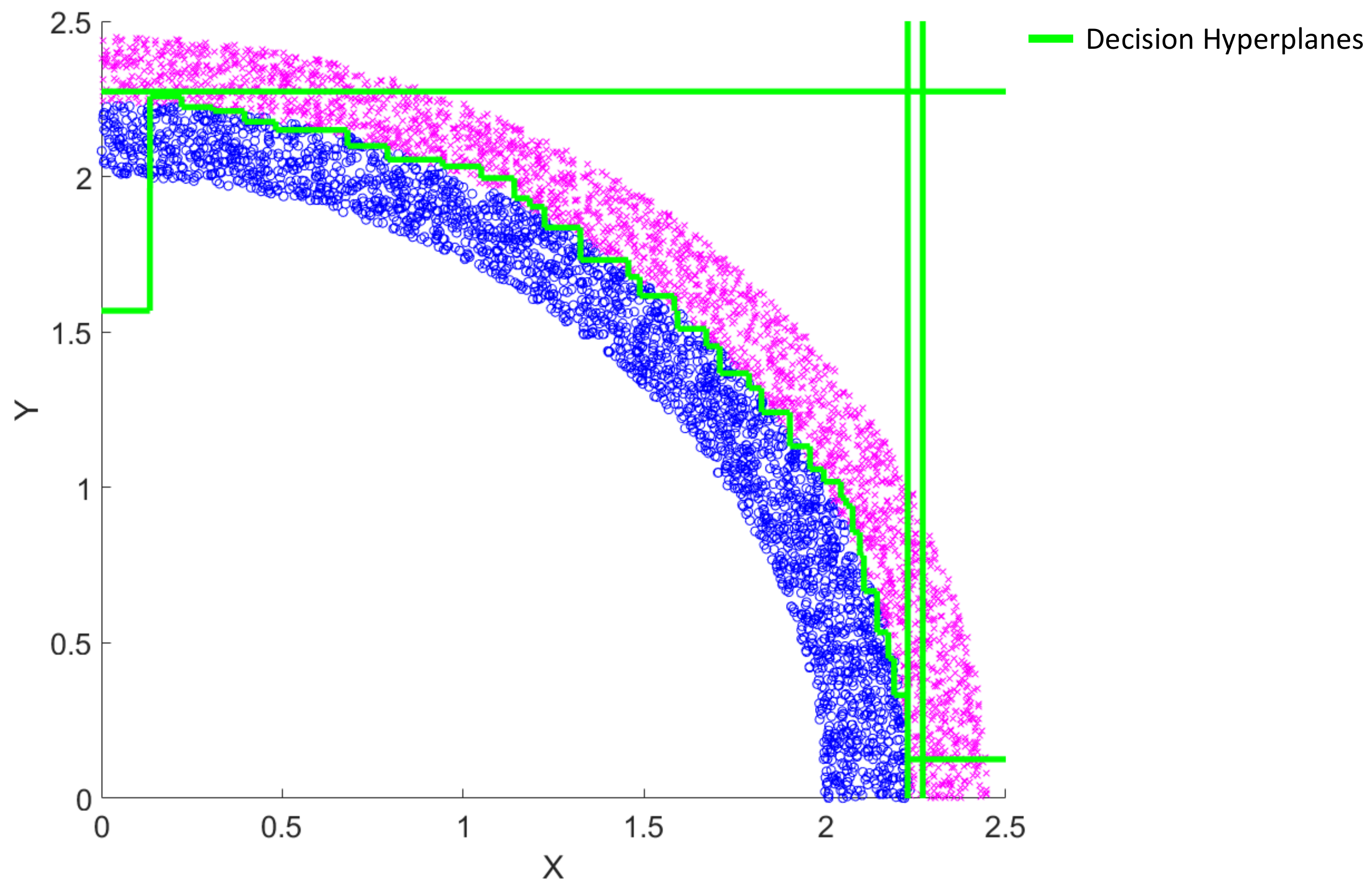}\label{fig:decisionBoundaries:C5}
}
\subfloat[Extended C-Net]{%
  \includegraphics[width=0.5\columnwidth]{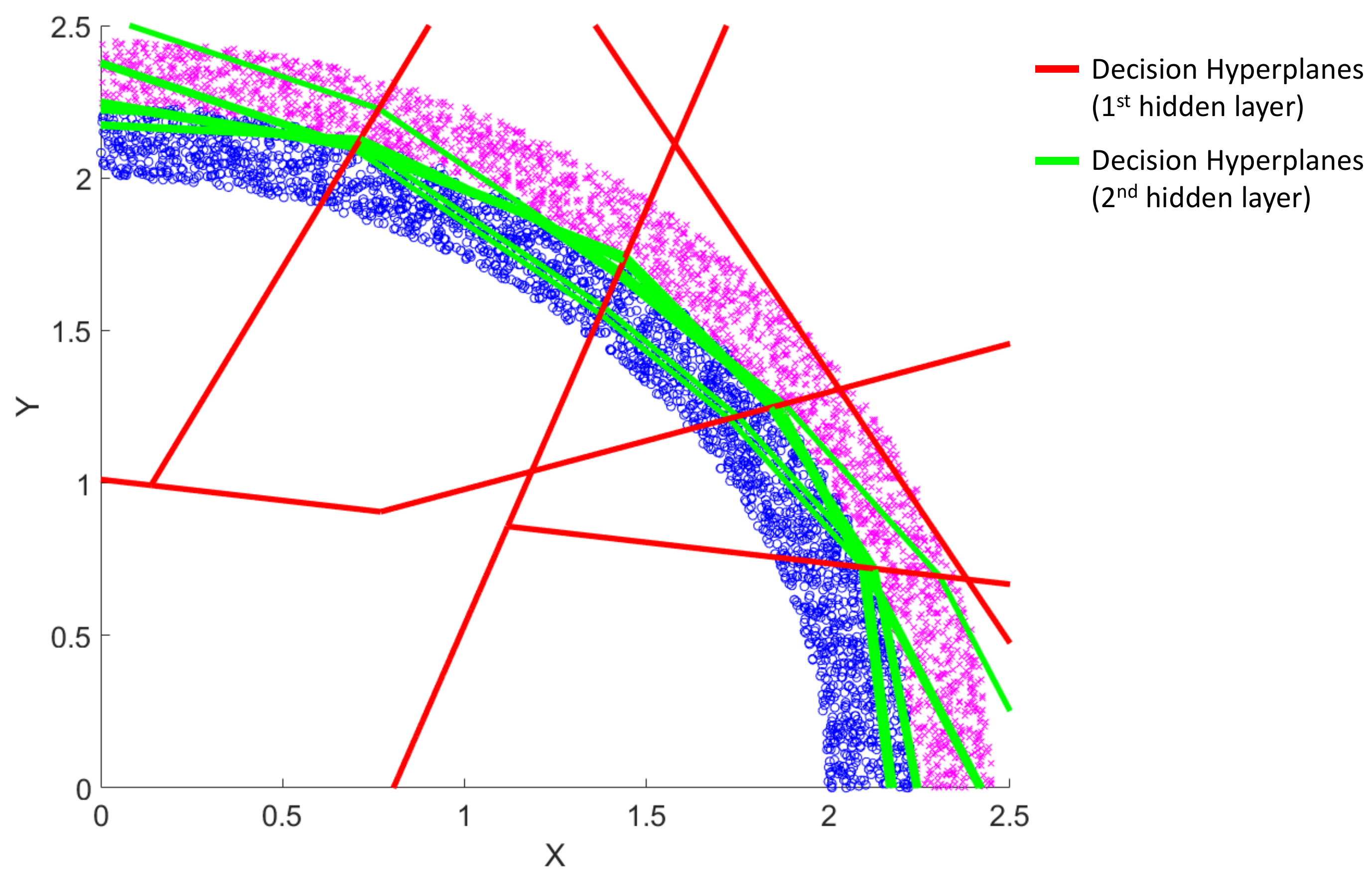}\label{fig:decisionBoundaries:CNet}
}

\subfloat[EC-DT]{%
  \includegraphics[width=0.5\columnwidth]{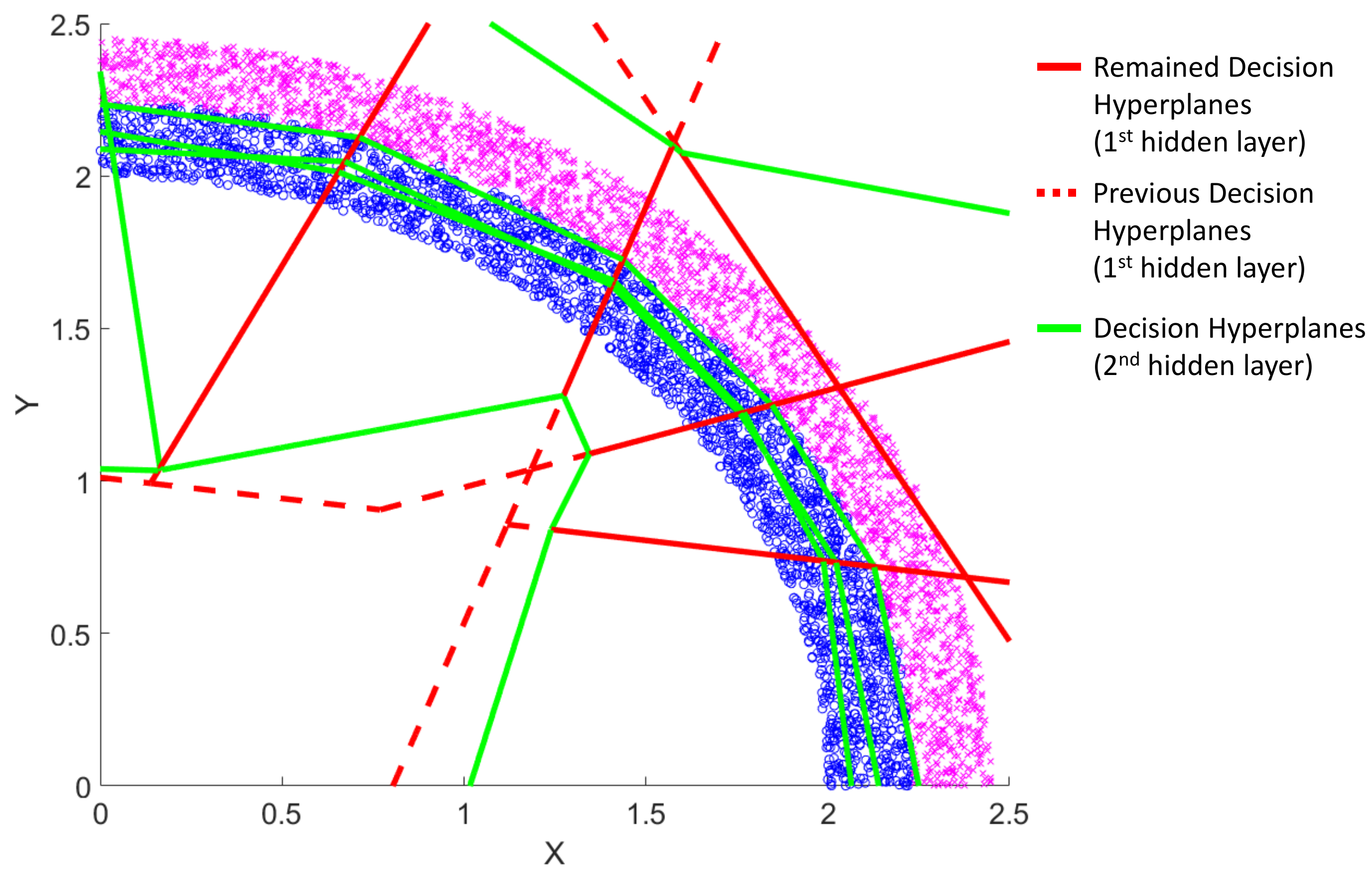}\label{fig:decisionBoundaries:EC-DT}
}

\caption{Decision hyperplanes extracted from pruned C5, Extended C-Net, and EC-DT transformed from a ANN model for P2 problem.}
\label{fig:decisionBoundaries}
\end{figure}

The synthetic problem P2 is representative for the set of highly nonlinear problems. To address this non-linearity, the simple C5 rule extraction algorithm has to generate a huge number of axis-parallel hyperplanes joining together to create a highly complex decision boundary as illustrated in Figure~\ref{fig:decisionBoundaries:C5}. On the contrary, Extended C-Net extracts the rules from ANN by considering the constraints for hidden nodes activation. Therefore, the set of rules from the first hidden layer tries to cluster the data space into sub-regions demonstrated by the area surrounded by the red lines (Figure~\ref{fig:decisionBoundaries:CNet}), while the second layer forms the constraints that are hyperplanes which at this stage directly separate two classes. In EC-DT, the tree representation of the true process of the ANN, the rules extracted from the first hidden layer resemblance the same clustering method as implemented by Extended C-Net. However, the set of rules extracted from the second hidden layer takes two roles at the same time where some nodes attempt to further shrink different data sub-regions and the others directly get involved in creating decision hyper-planes between two classes (see Figure~\ref{fig:decisionBoundaries:EC-DT}). The Extended C-Net, due to the employment of the similar rule structure at early layers of the EC-DT, exhibits a closer data processing to the ANN.

In summary, the complexity analysis sheds light on which rule extraction models one should choose. For problems where the classes are not linearly separable, or highly complicatedly distributed, one should employ multivariate approaches like EC-DT and Extended C-Net for higher rule completeness and higher fidelity. For problems requiring a high level of interpretability or that are simple, one can make use of a simple model like C5.0.

\subsection{Interpretation of Rules}\label{section-6.4}
In this paper, tree size is a compactness metric to analyze the global interpretability of the methodologies in the previous subsection. In this section, the instance-based interpretation of a rule is considered so that we can have a comprehensive view of how the explanation of a neural network as a black-box model can be generated.

\subsubsection{Case study: skin segmentation problem}
Given an instance in a dataset, an explanation can be generated from constraints which are included from an activated rule that the instance falls under. We provide an example of a rule so that we can evaluate the complexity and correctness of the constraints. We analyze the decision boundaries for the \emph{skin segmentation} dataset, which are a collection of samples extracted randomly from RGB images in FERET and PAL databases. These databases contain a variety of images of people with different characteristics such as age, race, and genders. Given an area in an RGB image, with three attributes of Green (G), Red (R), and Blue (B) values ranging between 0 and 255, one rule to identify that this area is human skin from a C5 tree can be found below:

\begin{quote}
\footnotesize
\textbf{IF}:
\begin{itemize}
    \item $B > 92$
    \item $G \leq 157$
    \item $R > 231$
\end{itemize}
\textbf{THEN}: This is \emph{skin}
\end{quote}

In case of Extended C-Net the constructed explanation for one rule can be displayed as below:
\begin{quote}
\footnotesize
\textbf{IF}:
\begin{itemize}
    \item $(-0.47*B) + (1.51*G) + (0.04*R) > -5.23$
    \item $(0.28*B) + (0.35*G) + (-0.47*R) > -4.00$
    \item $(0.69*B) + (-0.49*G) + (0.24*R) > -5.68$
    \item $(0.30*B) + (-0.66*G) + (0.32*R) > 4.22$
    \item $(0.53*B) + (-0.18*G) + (-0.29*R) > -10.14$
    \item $(2.75*B) + (-1.82*G) + (-0.81*R) > -5.37$
\end{itemize}
\textbf{THEN}: This is \emph{skin}
\end{quote}

Meanwhile, EC-DT shows a more complex explanation in exchange for the highest fidelity with the highest number of constraints:
\begin{quote}
\footnotesize
\textbf{IF}:
\begin{itemize}
    \item $(-0.47*B) + (1.51*G) + (0.04*R) > -5.23$
    \item $(0.28*B) + (0.35*G) + (-0.47*R) > -4.00$
    \item $(0.69*B) + (-0.49*G) + (0.24*R) > -5.68$
    \item $(0.30*B) + (-0.66*G) + (0.32*R) > 4.22$
    \item $(0.53*B) + (-0.18*G) + (-0.29*R) > -10.14$
    \item $(-0.32*B) + (0.50*G) + (-0.10*R) > 11.20$
    \item $(-0.38*B) + (0.56*G) + (-0.49*R) \leq -1.52$
    \item $(-0.64*B) + (0.27*G) + (0.29*R) > 7.15$
    \item $(-0.03*B) + (4.52*G) + (-4.14*R) \leq -36.45$
    \item $(2.75*B) + (-1.82*G) + (-0.81*R) \leq -7.76$
    \item $(3.32*B) + (-1.80*G) + (-1.27*R) + (46.98457) > 0$
\end{itemize}
\textbf{THEN}: This is \emph{skin}
\end{quote}

In the cases of Extended C-Net and EC-DT, the rules involve a combination between all attributes. The attributes with higher weights have more influence on the final decisions than those with lower weights. The positive/negative signs of the weights emphasize the contribution of the attributes towards positive or negative classes in the problem.

Despite the fact that the exhibition of rules extracted from Extended C-Net and EC-DT trees provides a more complete explanation of the decision, the complexity from the number and structure of the constraints within the rules are extremely higher than the C5 constraints. The rule as a result becomes less interpretable. Nevertheless, a way forward on this issue might come from an additional technique to transform the mode of explanation depending on problem. 

\begin{figure}[!ht]
\centering
    \subfloat[C5]{\includegraphics[width=0.35\linewidth]{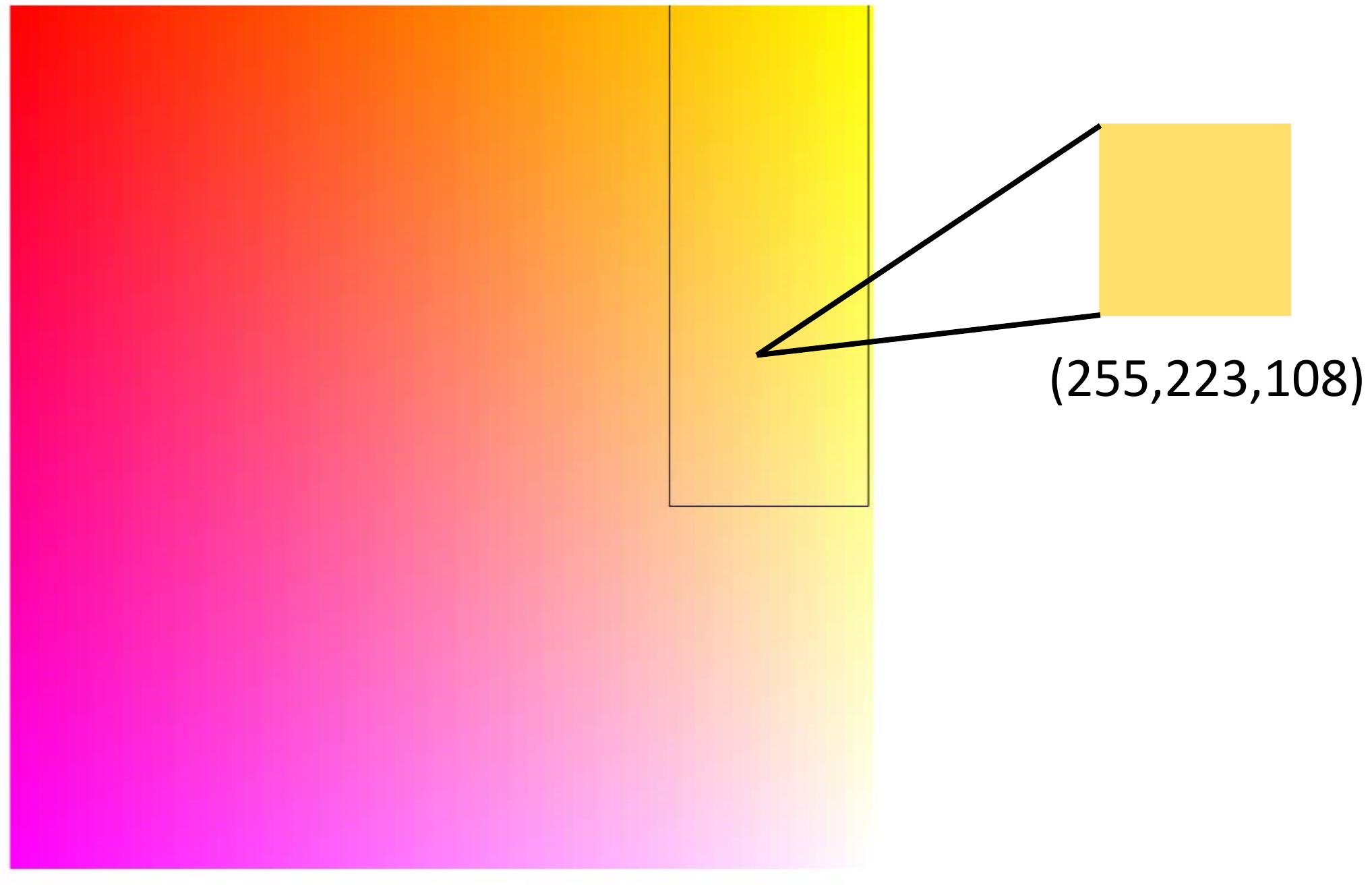}}
    \subfloat[Extended C-Net]{\includegraphics[width=0.35\linewidth]{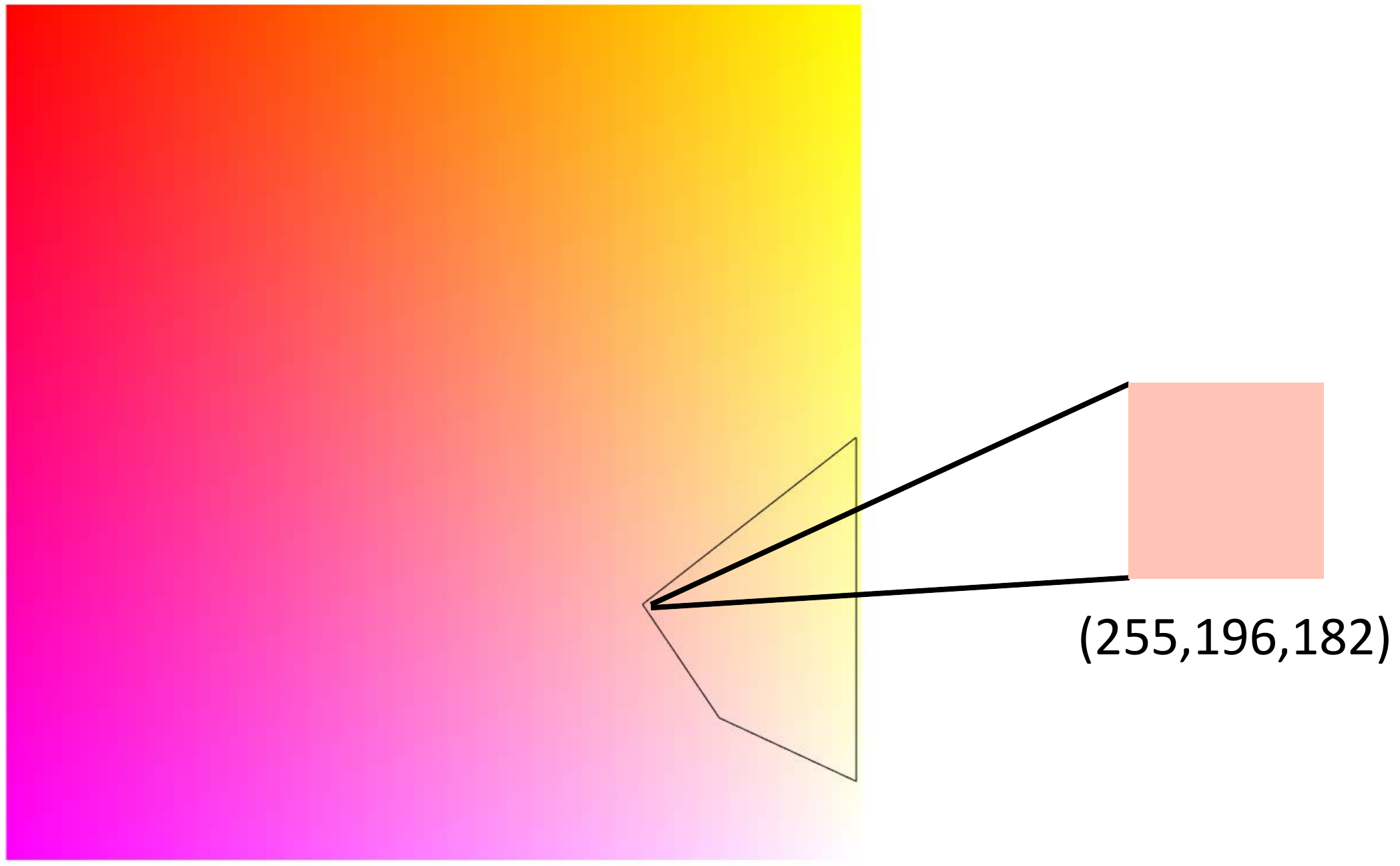}}
    \subfloat[EC-DT]{\includegraphics[width=0.35\linewidth]{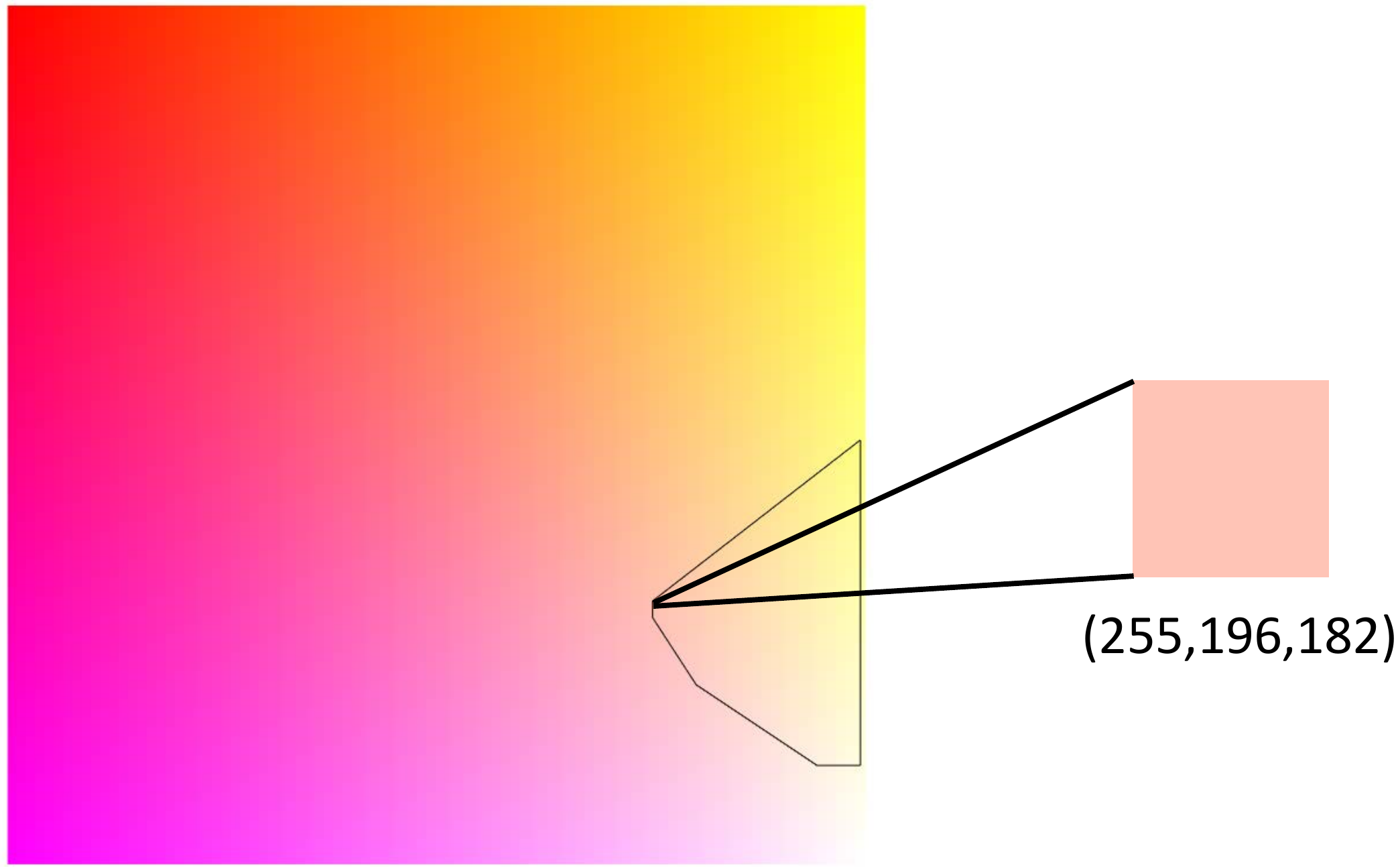}}
    \caption{The RGB colormaps with bounded regions representing corresponding rules from different tree models.}
\label{fig:skin_RGBmap}
\end{figure}

In the \emph{skin segmentation} problem, one might find it more appropriate to transform the rule constraints into visualization of the RGB colour maps and a sample of data needs to be explained can be also projected in the map. For example, given the red value $R=255$, we can construct the colour map for the rules above as illustrated in Figure~\ref{fig:skin_RGBmap}.

With this type of visualization, it is also interesting to note that the decision regions created by rules from C5 are axis-parallel rectangles which are simple, but less precise than the polygons and the shrunk polygons created by Extended C-Net and EC-DT respectively.

In previous literature regarding skin segmentation from images, a branch in computer vision, many studies~\cite{kovac2003human,vezhnevets2003survey} have introduced the explicitly defined skin region approach, which defines fixed conditions for RGB or YUV color ranges, to discriminate between skin and non-skin areas. While this method is simple, fast, and highly interpretable, it faces a significant challenge in achieving a competitive level of fidelity. Nevertheless, the method is still reliable to use as an initial screening method for skin detection~\cite{li2010face}. Many machine learning methods, including ANNs, have been used to enhance the detection rate in this problem. However, their interpretability is lower than that of the simple explicitly defined skin region approach. The translation of the rules extracted from a ANN into a visualizable representation might improve the transparency of the mappings. Our transformation of rules into colormaps shows similar utility to the visualization of color ranges in the literature~\cite{naji2012skin}.

\subsubsection{Case study: shepherding problems}
\begin{figure}[!ht]
\centering
    \subfloat[]{\includegraphics[width=0.6\linewidth]{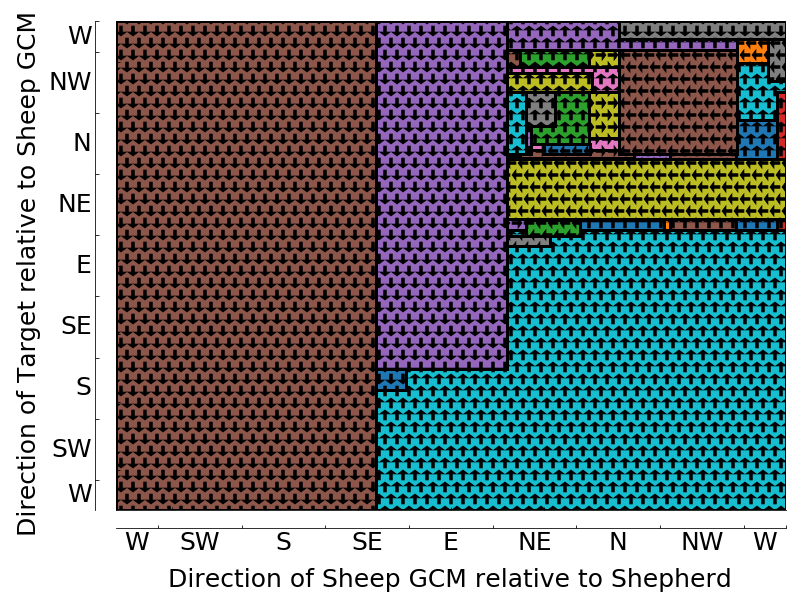}}
    \subfloat[]{\includegraphics[width=0.6\linewidth]{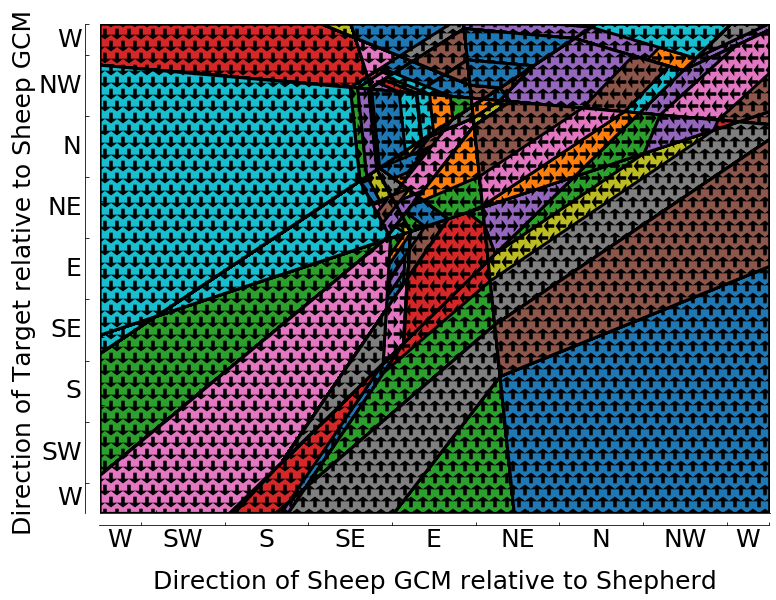}}\\
    \subfloat[]{\includegraphics[width=1.2\linewidth]{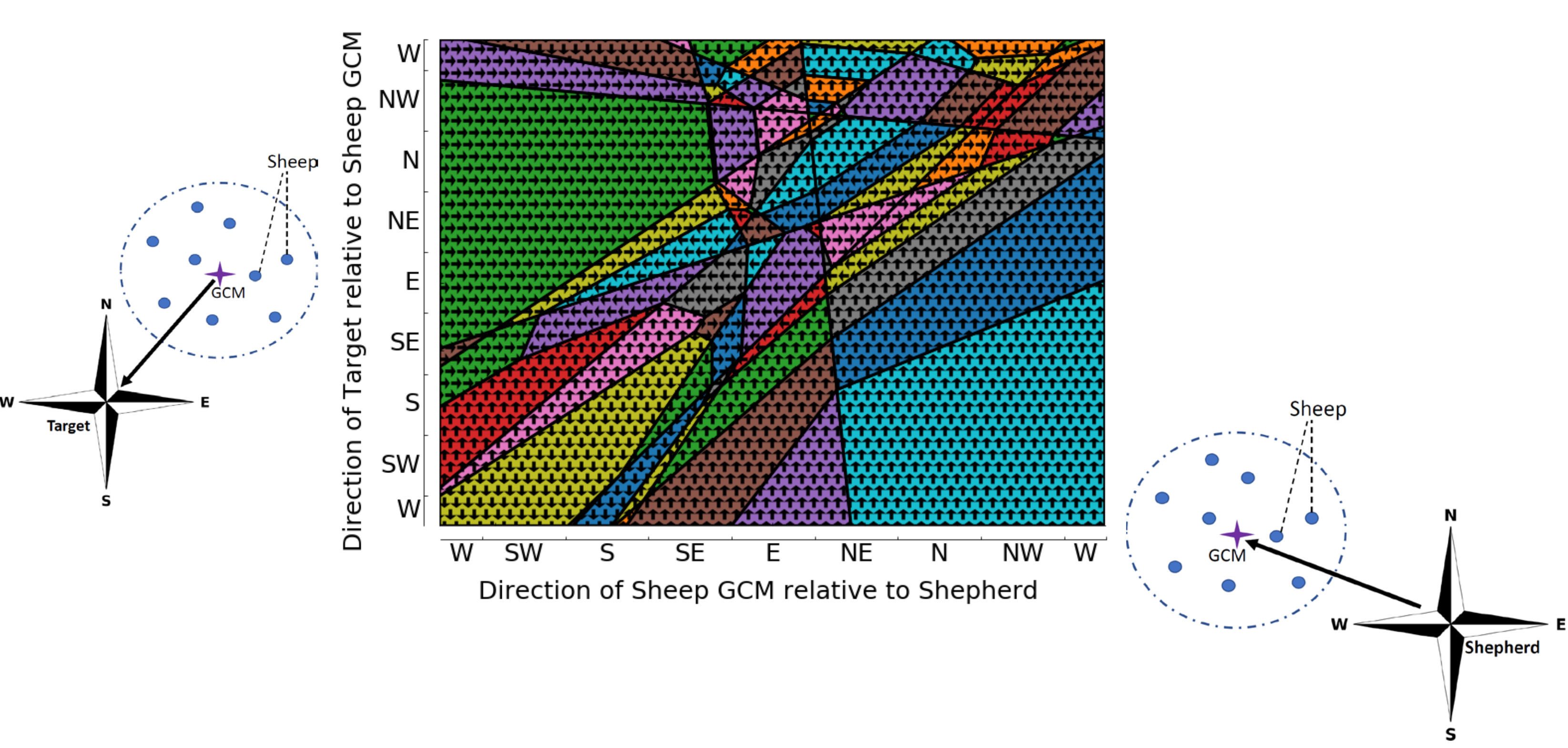}}
    \caption{The decision polytopes characterizing the driving behaviour of the shepherd with (a) C5, (b) Extended C-Net, and (b) EC-DT model. The distance between the shepherd and sheep\textquoteright \ GCM, and the distance between the sheep\textquoteright \ GCM and the target are 50 and 100 metres respectively. The directions of arrows in the polytopes are output directions of the shepherd.}
\label{fig:shepherding_driving}
\end{figure}

\begin{figure}[!htbp]
\centering
    \subfloat[]{\includegraphics[width=0.6\linewidth]{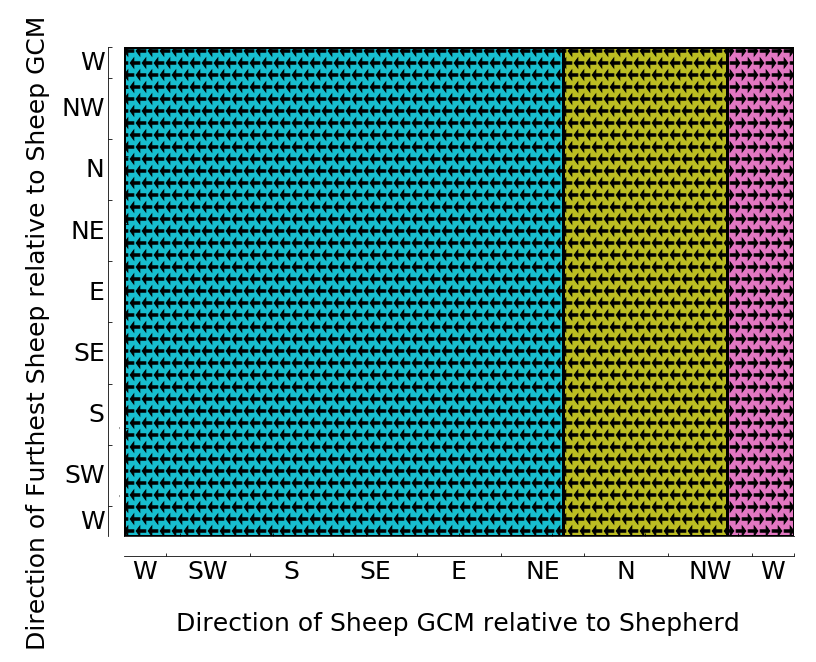}}
    \subfloat[]{\includegraphics[width=0.6\linewidth]{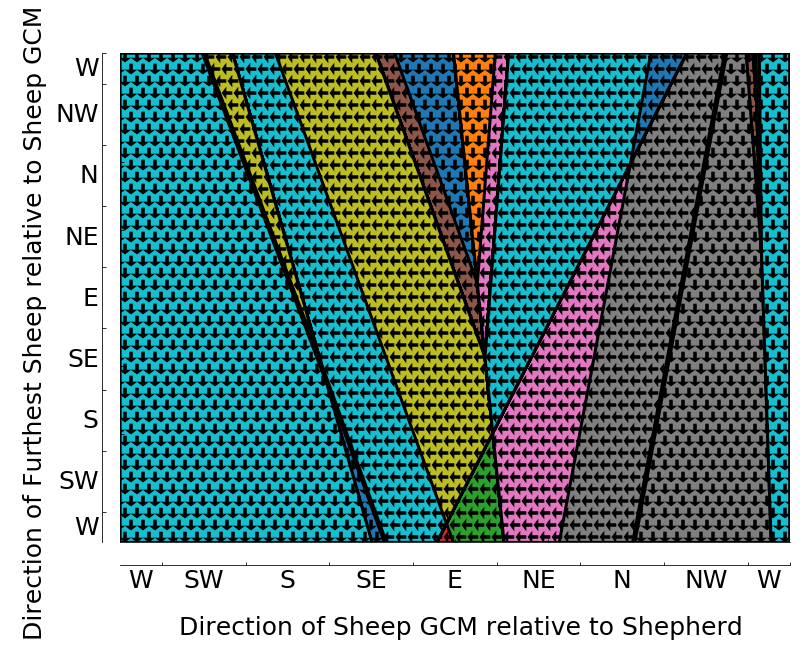}}\\
    \subfloat[]{\includegraphics[width=1.2\linewidth]{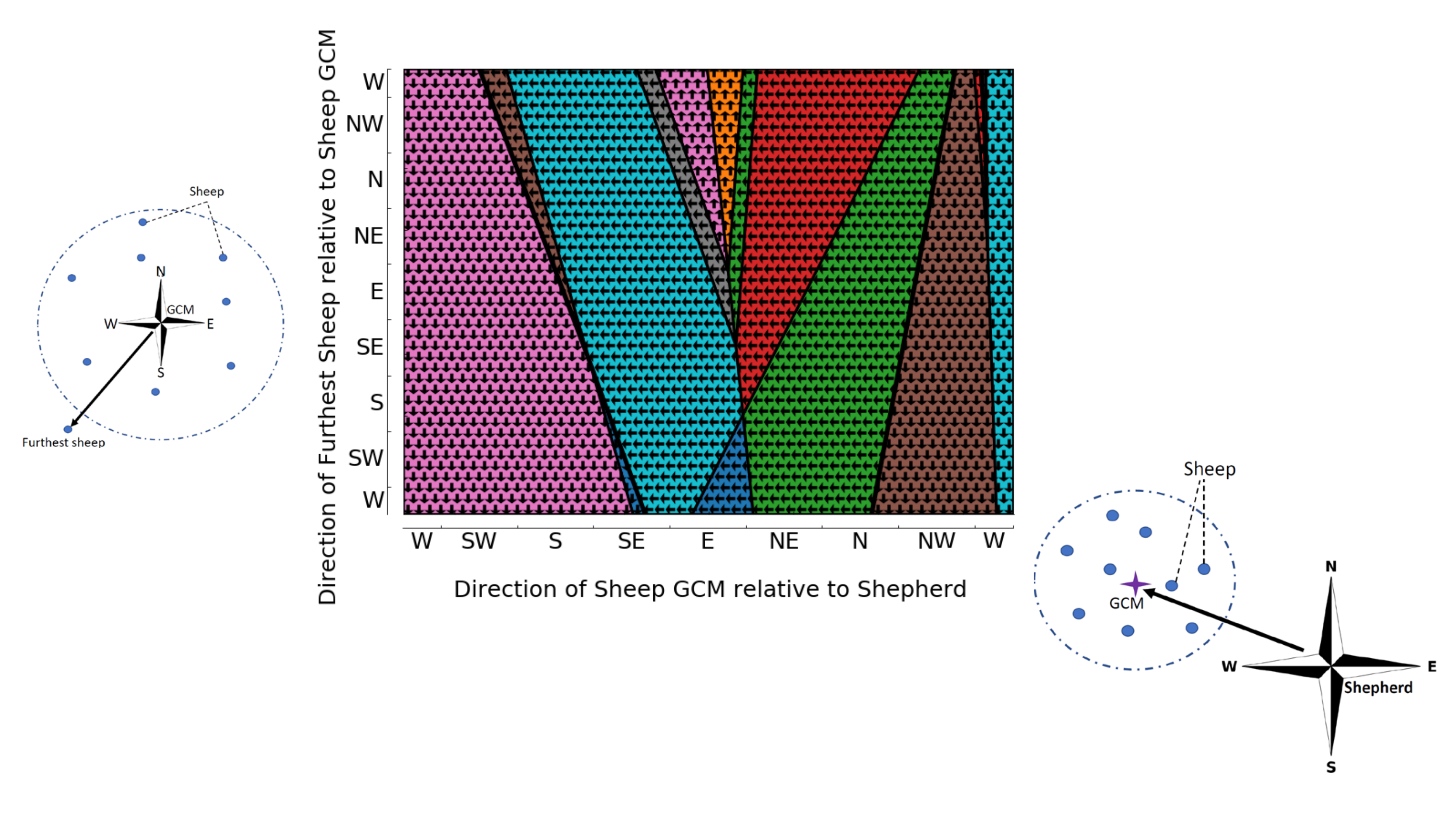}}
    \caption{The decision polytopes characterizing the collecting behaviour of the shepherd with (a) C5, (b) Extended C-Net, and (b) EC-DT model. The distance between the shepherd and sheep\textquoteright \ GCM, and the distance between the sheep\textquoteright \ GCM and the furthest sheep are 40 and 40 metres respectively. The directions of arrows in the polytopes are output directions of the shepherd.}
\label{fig:shepherding_collecting}
\end{figure}

As another example, we convert the rules extracted from C5, Extended C-Net and EC-DT trees into a visual explanation of the decision-making processes of the neural network as illustrated in Figure~\ref{fig:shepherding_driving} and Figure~\ref{fig:shepherding_collecting} corresponding to \textit{driving} and \textit{collecting} behaviours in \textit{shepherding}. The direction of the movement force of the shepherd is originally modelled as a function of the direction between the sheep\textquoteright \ global centre of mass (GCM) and the shepherd, and the direction between the target/furthest sheep and the sheep\textquoteright \ GCM in previous literature~\cite{strombom2014solving}. The visualization provides a map of decisions with regards to the input states of relative directions of interest.

Similar to the \emph{skin detection} problem, the decision regions created by rules from C5 are also axis-parallel rectangles. This offers a relatively easy way to illustrate general relationships between variables that result in the outputs of the model. Nevertheless, this visualization of the model behaviour is much less precise than the oblique polygons created by Extended C-Net and EC-DT. Figure~\ref{fig:shepherding_driving} illustrates the decision polytopes of shepherd\textquoteright s driving behaviour models extracted from the driving neural network in regards with two dimensions of the input states, which are the direction of the sheep GCM relative to the shepherd agent (x-axis) and the direction of the target in relative to the sheep GCM (y-axis), both are represented by angles (in degree) of corresponding directional vectors in the Unit Circle. The driving behaviours of the shepherd represented by the outputs of the polytopes generated by Extended C-Net and EC-DT rules are highly similar to each other. However, the EC-DT rules are better learned evidenced by the higher number of divisions of the region characterised by coordinates $x \in (-150,-150)$ and $y \in (-150,-50)$ in the state space where Extended C-Net rule model, in fact, overlooks. The observed variations above contribute to the differences of the fidelity of the models to the original neural network.

\begin{figure}[!ht]
\centering
\includegraphics[width=\linewidth]{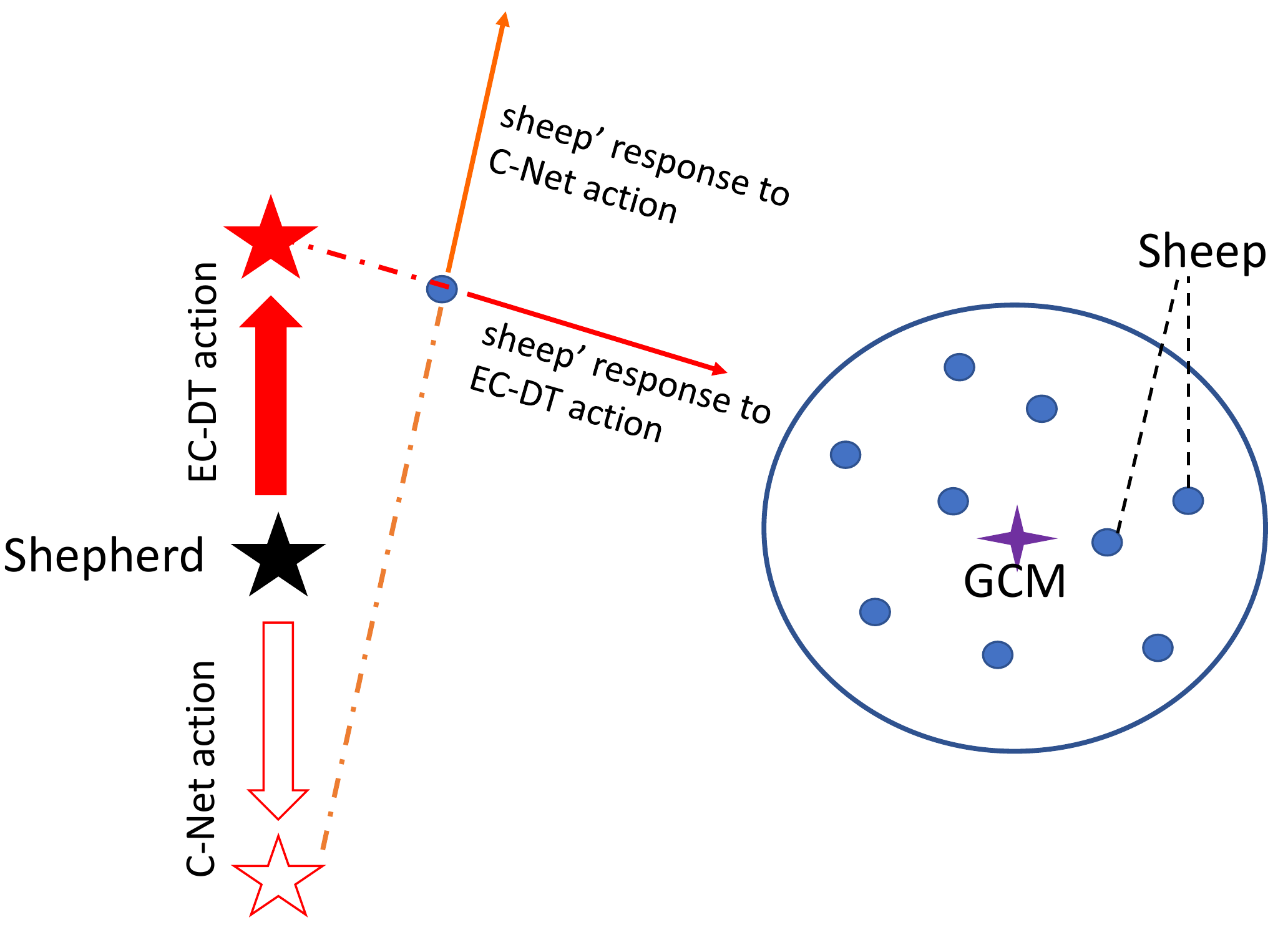}
\caption{The actions of the shepherd according to Extended C-Net and EC-DT models and the reaction of the sheep in collecting task.}
\label{fig:comparing_collecting}
\end{figure}

The collecting behaviours of the shepherding problem using the collecting network are modelled by three rule extraction algorithms. These decision polytopes corresponding to the behaviours based on two input dimensions are visualized in Figure~\ref{fig:shepherding_collecting}. In this visualization, we can observe the same interpretability and precision characteristics of the rule models as mentioned with rule models for driving behaviour above. The behaviours of the Extended C-Net and EC-DT models are similar to each other except a small region at coordinates $x \in (-20,10)$ (the sheep GCM is around north of the shepherd) and $y \in (50,180)$ (the fursthest sheep is between northeast and west relative to the sheep GCM). Figure~\ref{fig:comparing_collecting} demonstrates an example state of such situation. The optimal behaviour for this case is to move \emph{north} so that the positions of the shepherd, the furthest sheep, and the sheep GCM are alligned, which helps maximise the effectiveness of the sheep collection behaviour. The EC-DT shows a higher fidelity to the original neural network and provides a better performance by choosing to move \emph{north} in this situation while the Extended C-Net shows a less effective solution by selecting \emph{south} direction. Such lower fidelity of the Extended C-Net rules accounts for the poorer performance represented by the success rate and the number of steps to complete the task in the shepherding problem compared to that of EC-DT model despite its better rule compactness.

\subsubsection{Section remarks}

Our transformation of rules into maps of regions and vectors helps alleviate the lack of transparency of the mathematical forms of rules extracted from the ANNs. It also delivers a mode of representation, where not only the fidelity is maintained but higher complexity of the rules set from Extended C-Net or EC-DT are lessened to the same level of the one from C5.

The algorithms and analyses described in this paper may help practitioners to design a framework to analyze the complexity of the problems and utilize the rule extracted from the ANNs and customized visualization for investigating and trouble-shooting decision-making processes of models. Doshi-Velez  and  Kim~\cite{doshi2017towards}  propose  different sets of experiments for the evaluation of interpretability including application-based (evaluation directly on the applications), human-based (simple experiments with human evaluation), and functionality-based evaluation (experiments to test functionalities with no human), each of which rests on a different position of the spectrum for interpretability assessment. Our work is based on the setting of the functionality-based evaluation where we focus on assessing a rule-based interpretation of ANNs in terms of \textit{correctness} and \textit{compactness} of the interpretable models, which are objective metrics, as suggested by~\cite{silva2018towards}. Nevertheless, the interpretability of models is also subjectively dependent on different users. It is necessary to extend this research to the human and application-based levels in the future to capture the subjective assessment of interpretability.

\subsection{Computational Complexity}\label{section-6.5}

In this section, we analyze the computational time for different rule extraction methods. The computational time is an important criterion to evaluate if an algorithm is scalable for real-world problems, which might include a large number of data points.

\begin{figure}[!ht]
    \centering
    \includegraphics[width=\linewidth]{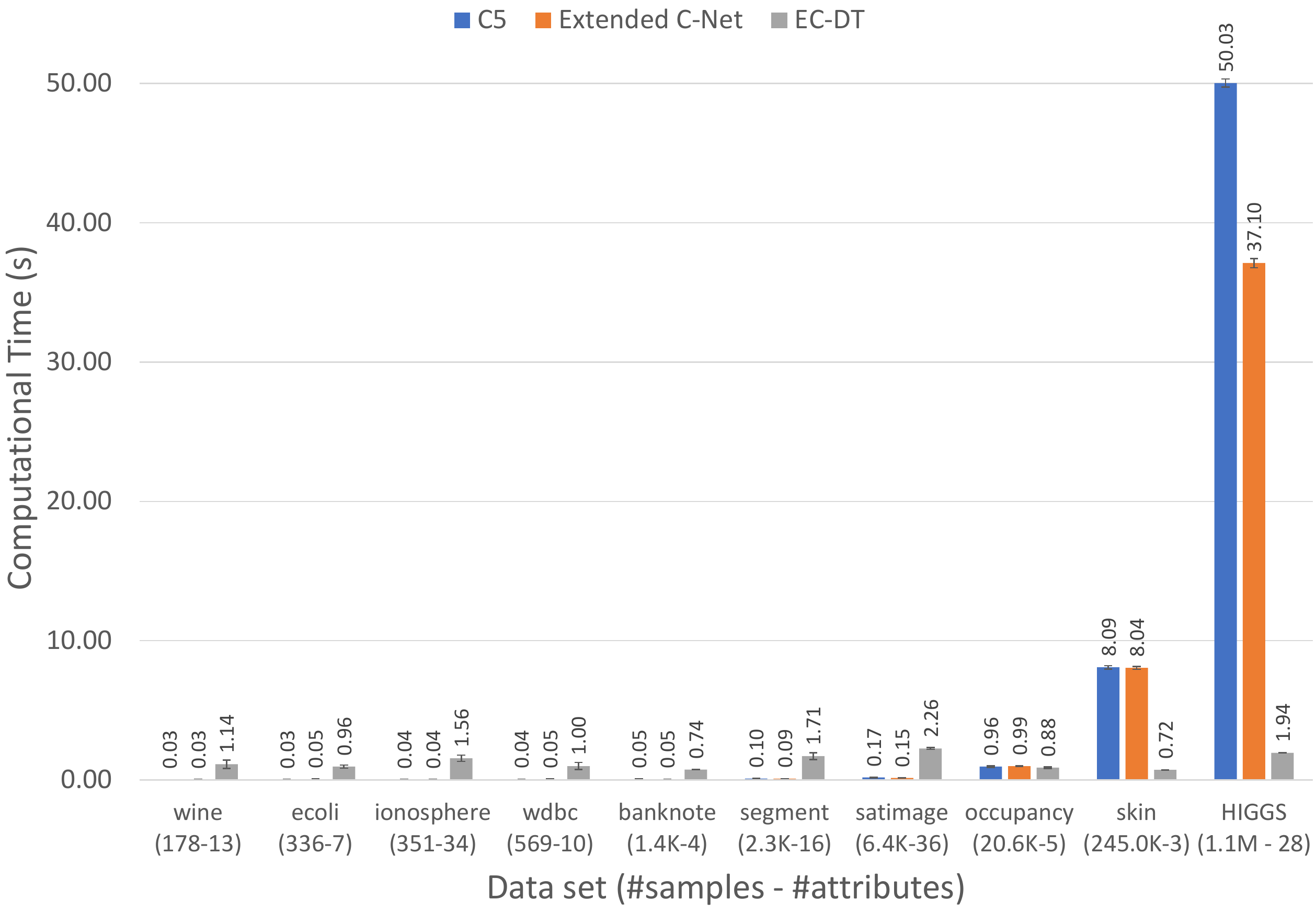}
    \caption{Computational time for generating global rule sets with C5, Extended C-Net, and EC-DT algorithms for different data sets.}
    \label{fig:computation-by-samples}
\end{figure}

Firstly, we investigate the scalability of different rule extraction algorithms with an increase in the number of samples. For this analysis, the computational time for generating global rule sets extracted with C5, Extended C-Net, and EC-DT algorithms are considered. Due to the fact that LIME, s-LIME, and Anchor are local interpretation algorithms, they do not generate the global set of rules from the beginning like the decision tree methods. Therefore LIME, s-LIME, and Anchor are not included in this comparison. Figure~\ref{fig:computation-by-samples} shows the computational time of different rule extraction algorithms on multiple data sets. The horizontal axis lists the data sets in the ascending order of number of samples and data attributes. For C5 and Extended C-Net algorithms, the computation speeds are significantly faster than EC-DT on data sets with less than 21000 data. As C5 and Extended C-Net are data-driven and hybrid methods respectively, their computational complexity strongly depends on the number of samples and data attributes. For large data sets like ``skin'' and ``HIGGS'' where the number of samples can reach hundreds of thousands or a few millions, the computational time becomes significantly larger than EC-DT method. The computational time of Extended C-Net in these cases are lower than that of the C5 algorithm. The computational time of EC-DT does not increase with an increasing number of samples. The computational time, however, sightly fluctuates depending on the number of data attributes in the data sets.

Even though the computational time of EC-DT is less likely to be influenced by the number of samples in a data set, it is strongly associated with the number of hidden nodes in the ANNs. An experiment is carried out on multiple ANNs trained on ``HIGGS'' data sets where the number of hidden nodes varies from 10-30. The computational time for EC-DT to generate an entire tree from ANNs with 10, 20, and 30 hidden nodes exponentially increases from $1.94s$ to $2.04\times 10^3$ and $2.06\times10^6$ seconds respectively. This is due to the fact the number of combinations of node activation configuration in a neural network exponentially increases with the number of hidden nodes, which makes EC-DT impractical for large ANNs. One has to sacrifice the fidelity of the interpretation of the neural network in exchange for usefulness of a model when interpreting a large neural network.

\begin{figure}[!ht]
    \centering
    \includegraphics[width=\linewidth]{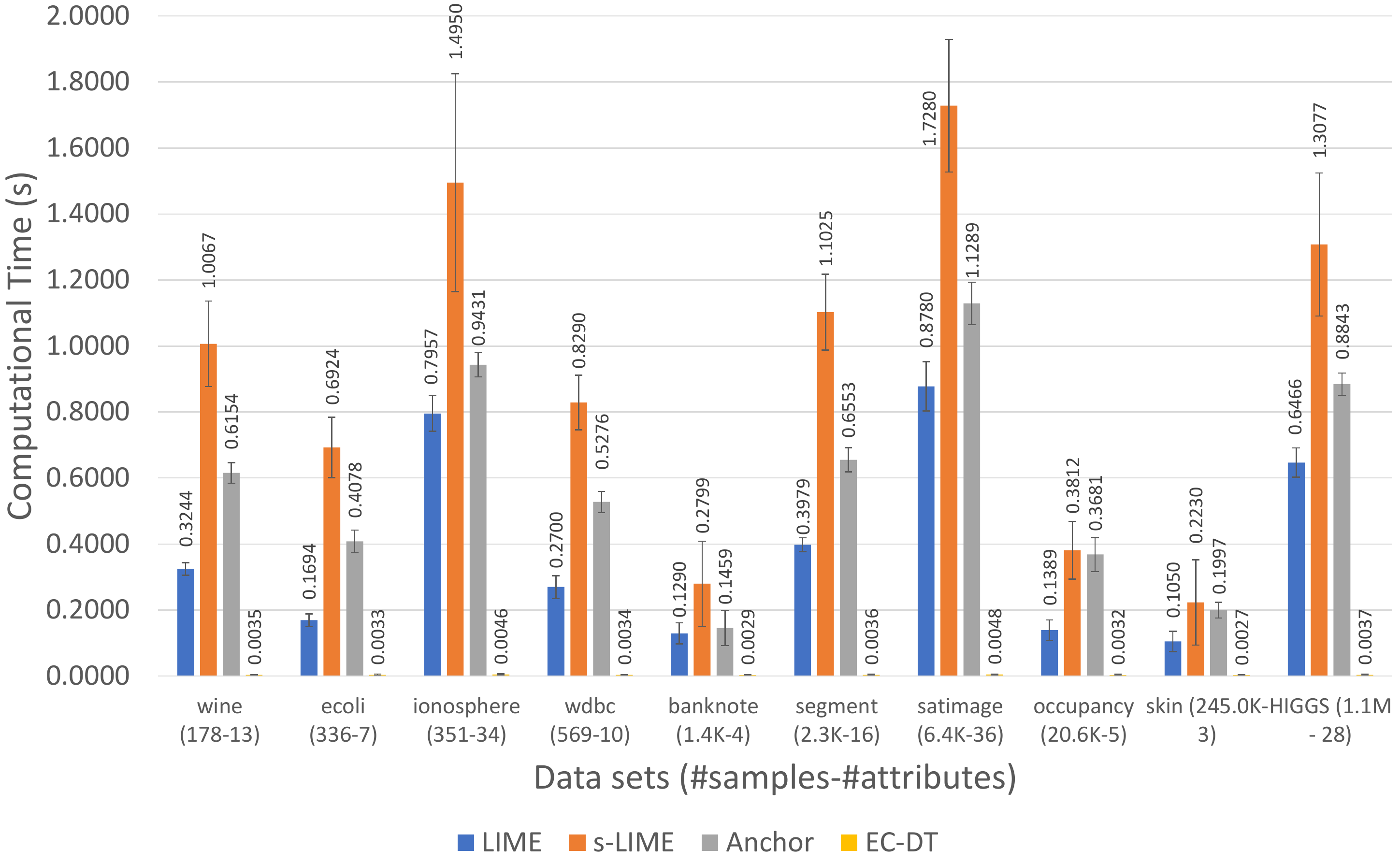}
    \caption{Computational time for generating local interpretation with LIME, s-LIME, Anchor, and EC-DT algorithms for different data sets.}
    \label{fig:local-computation-time}
\end{figure}

Although the lack of usefulness of EC-DT algorithm to transform an entire large neural network into a global set of rules is a significant limitation, one can still make use of it for ``on-demand'' interpretation of the neural network models. The EC-DT algorithm originally extracts the entire rule set which is an interpretable exact representation of the neural network. Instead of expensively computing all possible rules in the rule representation, one can compute a local interpretation of the neural network similar to LIME and Anchor algorithms. For a given input-output pair, the local EC-DT can extract corresponding rule based on the activation of the nodes in the neural network. This approach is less expensive and can still preserve a same level of fidelity of the representation. To evaluate the usefulness of this approach, we compare the mean computational time for generating a local interpretation with LIME, s-LIME, Anchor, and local EC-DT methods. Figure~\ref{fig:local-computation-time} illustrates the computational time for different local interpretation methods on multiple data sets. EC-DT achieves an average of 0.0035s for generating a local interpretation, which is significantly lower than all other methods like LIME, s-LIME and Anchor. LIME, s-LIME and Anchor algorithms use the input samples to generate synthetic neighbour samples to estimate the local rule, which costs more time than decompositional method like EC-DT.

\section{Conclusion}\label{section-7}

In this paper, we propose two novel multivariate decision tree frameworks which can generate interpretable rules for explaining the operation of ANNs. The first framework is a modification of Extended C-Net algorithm into a Extended C-Net which can learn the relationship between the last layer of a ANN and the output to back-project and extract the multivariate constraints on input as a set of highly accurate rules. The second is an algorithm called EC-DT that can directly translate the ANN layer-wise and build the set of rules with 100$\%$ fidelity to the ANN. 

EC-DT offers the best fidelity when it perfectly preserves the performance of the ANNs, but with the cost of a larger number of rules and number of constraints in each rule. It is understandable as the high number of rules are to capture the generalization that the ANN offers. Compared to a traditional approach of generating trees with a baseline C5 algorithm or a local explainer such as Anchor, Extended C-Net in general can better maintain the fidelity of the ANN while achieving the most compact set of rules. 

However, the use of simple versus complex models results in the trade-off between the simplicity and interpretability against the fidelity and precision. To decide on which model to use, one should consider the complexity of the problem space. For linear-separable classification problems, a classic C5 and Anchor can achieve similar results to ANN with a very low and simple set of rules. In situations where it is important to have a high interpretability or high speed of generating the explanation, a local approximation algorithm like Anchor would be more efficient as this algorithm does not need to build the entire tree like what C5 does. For highly nonlinear problems where a large number of attributes are involved, EC-DT and Extended C-Net exhibit significantly higher fidelity than a simple C5 or Anchor. In general, in the situation where the priority is fidelity, the EC-DT can be used, while in cases where the balance between fidelity and interpretability is required, Extended C-Net is favored.  

The weakness of the more complex models such as Extended C-Net and EC-DT comes from the large number of multivariate constraints for each rule, where the form of C5 constraints is very simple. The plain display of the mathematical conditions as an explanation might lower the transparency. Therefore, a suitable transformation of the representation of rules to some explanation mode that reduces the number of dimensions can be employed to overcome the issue. The visualization of decision hyperplanes that is introduced in this paper in a specific problem of \emph{skin segmentation} and \emph{shepherding control task} are examples for an effective explanation interface for instance-based interpretation.

Another limitation of EC-DT algorithm is that it is only useful for small networks. The computational time for EC-DT to generate an entire set of rule, which preserves a high fidelity, increases exponentially with an increase in the number of nodes in the neural network. However, the EC-DT can still be useful when it is used as an local interpretation algorithm like LIME and Anchor. Local EC-DT can achieve the same fidelity level of individual interpretations with a much significantly lower computation cost compared to LIME, s-LIME, and Anchor.

In future work, the interpretability of rules extracted from our proposed EC-DT and Extended C-Net algorithms will be investigated on more problems. The final interpretability is a fusion between the objective metrics, such as the compactness and complexity of the representation (in our case, approximated by the number of rules and the number of constraints, respectively) and subjective metrics, which are not covered in this paper. Therefore the suitability of the extracted knowledge may contribute to new pieces of knowledge to different human experts, which would call for a subject-matter expert-evaluation of the extracted knowledge. 

\bibliography{references}
\end{document}